\RequirePackage{fix-cm}
\documentclass[smallextended]{svjour3}       
\smartqed  
\usepackage{graphicx}

\usepackage{tabularx}
\usepackage{caption} 
\usepackage{graphicx}
\usepackage{amsmath}
\usepackage{amssymb}
\usepackage{rotating}
\usepackage{subfigure}
\usepackage{multirow}
\usepackage[ruled,vlined]{algorithm2e}
\usepackage{url}
\usepackage[justification=justified,singlelinecheck=false]{caption}
\usepackage{colortbl}
\usepackage{enumitem}
\usepackage[table,xcdraw]{xcolor}
\usepackage{subfigure}
\usepackage{authblk}

\usepackage{geometry} 

\begin{document}

\title{ROBUST MODELING OF EPISTEMIC MENTAL STATES}


\titlerunning{Robust Modeling ...}        

\author[1]{AKMMahbubur Rahman}
\author[2]{ASM Iftekhar Anam}
\author[3]{Mohammed Yeasin}
\affil[1]{Independent University Bangladesh, Dhaka, Bangladesh\\   
Email: akmmrahman@iub.edu.bd, }
\affil[2]{University of Wisconsin Green Bay, Green Bay, Missouri, USA \\  Email: anami@uwgb.edu }
\affil[3]{University of Memphis, Memphis, Tennessee, USA  \\     Email: myeasin@memphis.edu }
\authorrunning{AKMM. Rahman et al.} 

\date{Received:  January 17, 2020; Accepted: May 26, 2020\\
Multimedia Tools and Application, Special Issue: Socio-Affective Technologies}

\maketitle
\begin{abstract}
This work identifies and advances some research challenges in the analysis of facial features and their temporal dynamics with epistemic mental states in dyadic conversations. Epistemic states are: Agreement, Concentration, Thoughtful, Certain, and Interest. In this paper, we  perform a number of statistical analyses and simulations to identify the relationship between facial features and epistemic states. Non-linear relations are found to be more prevalent, while temporal features derived from original facial features have demonstrated a strong correlation with intensity changes. Then, we propose a novel prediction framework that takes facial features and their nonlinear relation scores as input and predict different epistemic states in videos. The prediction of epistemic states is boosted when the classification of emotion changing regions such as rising, falling, or steady-state are incorporated with the temporal features. The proposed predictive models can predict the epistemic states with significantly improved accuracy: correlation coefficient (CoERR) for Agreement is $0.827$, for Concentration $0.901$, for Thoughtful $0.794$, for Certain $0.854$, and for Interest $0.913$. 
\end{abstract}


\section{Introduction} 
The natural social interaction is dominated by  complex mental states, with expressions of certainty, interest, concentration, easiness, and agreement \cite{Mahmoud2011}, \cite{palm2013towards}, \cite{shazia}.  These states can be illustrated as the feelings or emotions while people deal with the formalization of epistemological concepts such as reasoning, logic, and certainty \cite{baron2007mind}. These complex emotions  are commonly known as Epistemic Mental States (EMSs) \cite{semain}, \cite{Mahmoud2011}, \cite{baron2007mind}. Epistemic states are highlighted by Baron-Cohen et al. \cite{baron2007mind}, and have been viewed within the machine learning community as significant elements in social context \cite{Mahmoud2011}. \cite{Roberts_Tsai_Coan_2007}.  Additionally, the link between these EMSs and learning environment has been established by a number of researchers from the fields of psychology, education, and computer science \cite{craig2004emotions}, \cite{goleman1995emotional}. A large number of psychological research provide evidence how learners’ knowledge and goals drive their moods and emotions during learning new topics \cite{carterette1996handbook}, \cite{mandler1975mind}.  The papers \cite{parkinson1993making}, \cite{kort2001affective} also  suggest that the learners experience a variety of above-mentioned EMSs depending on the context, the amount of new information, and the relation of new information towards their learning goal.
%

This shift towards the analysis of epistemic states is undoubtedly necessary if the modern social technologies such as artificial tutoring agents, assistive devices, and social agents are expected to be affect-sensitive. The examples of such agents can range from tutoring agents \cite{d2012autotutor} \cite{azevedo2009metatutor} to social robots\cite{breazeal2003emotion},  from assistive agents  \cite{borg2015assistive} for  kids to those for people who are blind or visually impaired  \cite{rahman2017emoassist}\cite{mcdaniel2018tactile}. Most importantly, tutoring agents and social agents would be popular in this busy world where people can learn or share their feelings without any human instructor or friend. An affect-sensitive system is a very important part to build these kinds of socially active agents. Consequently, towards building an affect-sensitive technology, we need to understand the role of human facial behavior on epistemic states. This paper identifies and advances research challenges in the analysis of the facial features and their temporal dynamics in emotionally colored human-human interaction. 

In the context of automated analysis, the basic expressions (anger, sad, fear, surprise, contempt, disgust) are analyzed by a plethora of research \cite{Cheon2009}, \cite{Zeng2007}, \cite{sydneystudy07}, \cite{surveysppechchannel},  \cite{huang2007labeled} for dyadic context. However, the EMSs such as certainty, agreement, interest, concentration, and thoughtfulness are not analyzed in a systematic  way thoroughly and sufficiently. Unlike the basic expressions, people exhibit subtle and critical changes in their faces during these states in face to face natural conversation \cite{Mahmoud2011}. These subtle changes make the prediction of these states quite challenging.   Therefore, it is necessary to understand the complex interplay between facial features and these epistemic states. 
Recently, some personal assistants such as Meta-tutor \cite{azevedo2009metatutor} can be thought of as  the primitive versions of social agents. Social robots from MIT \cite{gordon2016affective} could be the great example of social partners except they are not fully affect-sensitive. Especially, the personalized robots suffer from the lack of a robust framework that can estimate epistemic states in dyadic conversation. Most of them use voice and back-channeling to get some basic expressions. As we discussed earlier, the basic expressions are mostly absent in the dyadic context. Hence these robots might be ineffective in the targeted context. Particularly in a dyadic context and learning environment, visual signals are the most important input channels for human observers to perceive emotional states. Because of their unobtrusiveness, facial behaviors and pose are widely tried for emotion estimation in the machine learning community. However, a robust affect sensitive system that can estimate epistemic states from visual signals is not developed yet. Hence, this paper is also targeting for a robust framework that can capture the relations between facial features and epistemic states to predict states.


\begin{figure*}[!t]
\centering
\includegraphics[width = 5 in]{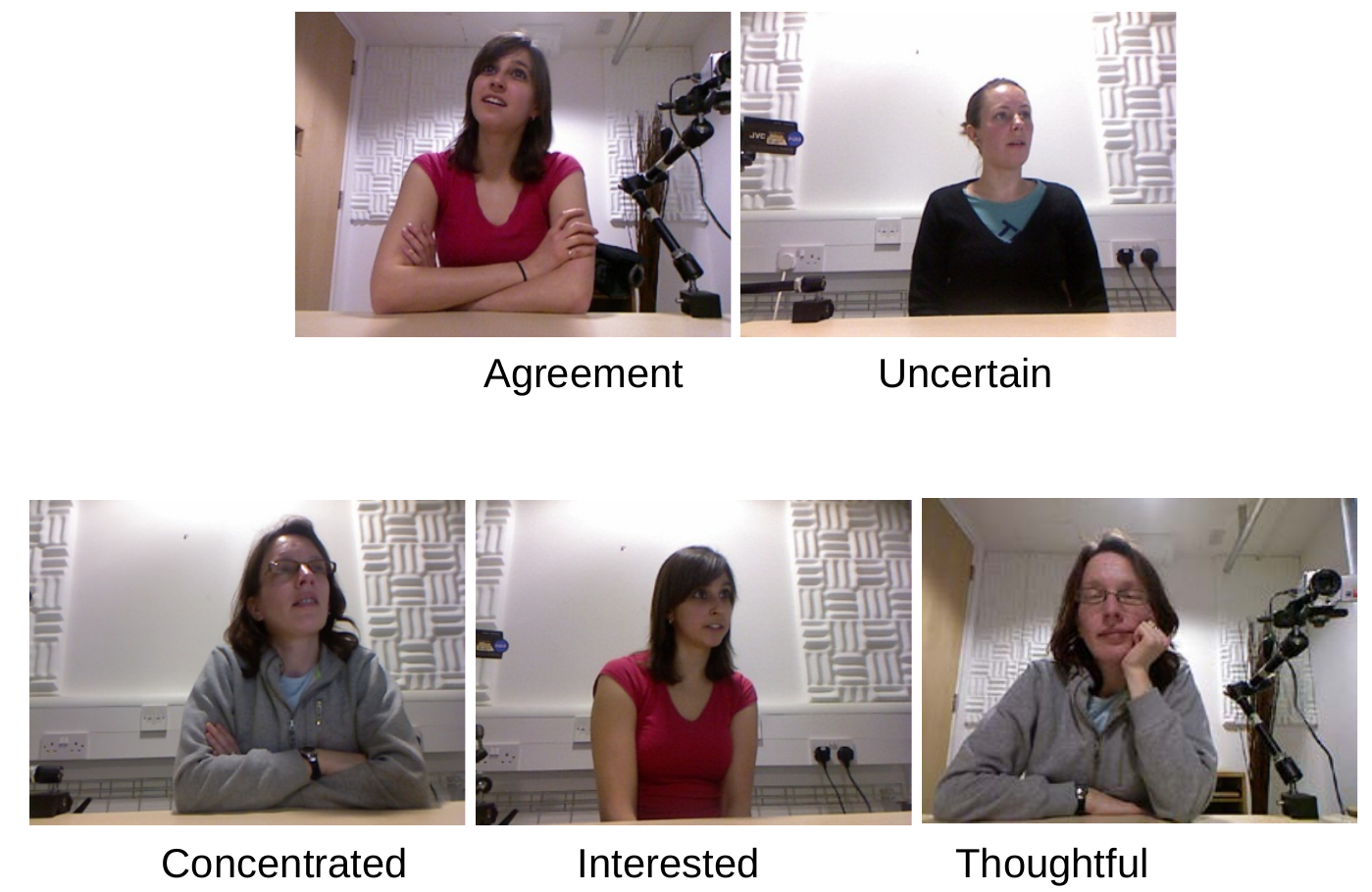}
\caption{Examples of five Epistemic States related to dyadic conversation \cite{Mahmoud2011}, \cite{shazia}}
\label{ExamplesofEpistemicStates}
\end{figure*}


Moreover, the exact nature of the interplay of facial features and their temporal evolution with EMSs are  still unknown at various levels of granularity \cite{cohn_schemidt}. A number of works are reported in the last decade to estimate Epistemic Mental States (EMS) from facial expressions. Most of them rely on the simple correlations  between  action units(AUs) and EMSs. McDaniel et al. \cite{mcdaniel2007facial1} identify the significant AUs that are strongly correlated with boredom, confusion, delight, frustration, and neutral. They measure the relations by calculating the simple Pearson’s correlation coefficient. Additionally, in \cite{bosch2014s}, Nigel Bosch  et al. detect engagement, frustration, confusion, and boredom through the AUs while the AUs are identified by CERT \cite{littlewort2011computer}. However, the results are not satisfactory because of low recall and low precision rate. In \cite{bosch2016detecting}, authors propose a supervised machine learning model where the authors build feature vectors from AUs \& their intensities. Then, they use these feature vectors in different weka based machine learning algorithms to predict the EMSs. However, they have done experiments on very limited data.  Actually the facial features exhibit subtle changes when that person is going through the complex EMSs. Subtle changes in facial features do not activate the AUs. Therefore, it’s clear that EMSs estimation by using AU detection is not effective. Instead of AUs, quantifying relations between facial features and EMSs is essential to capture the subtle facial changes. 
Therefore, in order to build a robust and reliable detection of EMSs, a number of important research questions should be answered regarding the complex relations among facial features and epistemic states. The key goal of this paper is to understand the interplay between facial features and to capture the subtle variations and relationship for robust modeling of an epistemic state of mind by answering the research questions. The research questions are (but are not limited to):
\begin{enumerate}
\item How are the facial features related to epistemic mental states? What are the characteristics of these relations? (cf. Section \ref{sec:ComplexInterplay})
\item	What are the temporal features and their temporal dynamics related to the epistemic state of mind?  What levels of granularity do they interact that can be used for their automated recognition in real-time? (cf. Section \ref{sec:Temporalfeature})
\item	 Are epistemic states related to each other? If so, how are they related? (cf. Section \ref{sec:EmotionvsEmotion})
\item	Is the duration of epistemic states different from prototypical emotional episodes? If so, what is their relationship? (cf. Section \ref{sec:intensities})
\item	Is it possible to capture the non-linear relationship in an automated way to model EMSs(cf. Section \ref{sec:AutomatedModeling}).
\end{enumerate}
By the levels of granularity, we want to mean that we want to find the answer of the research question: how and in what rudimentary levels, the temporal features are related to epistemic states and their changes over time. The granularity also indicates what orders the temporal features correlate with Epistemic states. We found that the temporal features i.e.  First order as well as second order derived features have very strong relations (high Maximal Information Criterion - MIC scores) with the Epistemic states. Our experiments also suggest that in the first order and second order features should be coupled with original features to calculate the nonlinear relations. After we get the answers of the research questions mentioned above, we  build an automated prediction framework for estimation of the epistemic state with intensities. The framework is also proposed to set up a benchmark for the research community. In summary, our contributions are listed as follows:
\begin{enumerate}
\item We answer the above mentioned research questions after performing a number of experiments.
\item We extract strength of non-linear relations between facial features and EMSs in terms of Maximal Information Coefficient (MIC) scores. We use Maximal Information-based Non-parametric Exploration (MINE) framework proposed by \cite{reshef2011detecting} to calculate the MIC scores.
\item We propose a novel benchmark  intensity prediction model for particular EMSs where
\begin{enumerate}
\item We use strength (MIC scores)  of non-linear relations in the classification and regression i.e. we use MIC score-weighted facial features.
\item We use MIC score-weighted temporal features.
\item We utilize a region classifier as an intermediate step of EMS prediction.
\end{enumerate}
\end{enumerate}

The three Sections \ref{sec:overview}, \ref{sec:intensities},  and \ref{sec:Episodes} provides sequential processes that help us  answer the questions above. Then, we reported our observations and key findings followed by a proposition of a new framework  based on our observations. At the end, the paper reports the performance of the proposed framework.

\section{Research Context} \label{sec:relatedWorks}

 In this section, we describe the recent works that propose different methods to capture significant relations between facial features and EMSs. At first, we discuss the EMSs in different contexts. Then we describe psychological evidence that demonstrate the relations between EMSs and various kinds of facial behavior. After that, we discuss the existing research works that propose machine learning based frameworks for estimation of EMSs automatically from facial expressions. At last, we present the limitations of the existing works and then continue to our proposed approach. 

The researches   \cite{d2012autotutor}, \cite{d2010multimodal}, \cite{metatutor},  \cite{Graesser_mindand}, \cite{sydneystudy07} identify EMSs that are interpreted as emotional states during learning.  While learning a specific topic, students go over different kinds of EMSs such as frustration, confusion, curiosity, eureka, and boredom. Interestingly, these emotions  are also more likely present in dyadic context \cite{shazia} \cite{Mahmoud2011}, \cite{shazia}, \cite{Roberts_Tsai_Coan_2007}. 
The paper \cite{lanzini2013different} refers the Epistemic Mental States (EMS) as the emotional states that involve cognition, perception, and feelings. Specifically, in this paper,  the  authors try to explore how humans interpret these emotions. After their observations, they conclude that human perceive these emotions through  multi-modal  features channels such as visual cues, auditory features, and even transcripts of face-to-face interactions \cite{lanzini2013different}. 

\cite{knapp2013nonverbal} suggests, when we provide verbal information, we produce an indicative body/head movement to share the information. The simple head nod is a universal symbol of agreement in many parts of the world. On the other hand, involuntary body movement can also produce visual signals of EMS. Spontaneous head nods can be perceived as a sign of understanding during a conversation. Hence, the person whom I’m talking with is nodding, so probably she understands what I’m saying.  If there is no body movements/visual cues from my conversation partner, my thinking may not be meaningfully shared with my partner.

Facial features  are always visible and active during a face to face interaction. Particularly,  mouth, eyebrows, and various facial muscles can offer important information about people’s  EMSs  \cite{cohn2010advances}. Generally, AU1 (Inner Brow Raiser), AU4 (Brow Lowerer),  AU12 (Lip Corner Puller),  AU 25 (Lips Part), AU M59 (head nods up/down), and AU M60 (head shakes sideways)  are involved in facial expression when people are going through the EMSs \cite{stratou2017refactoring}. Specifically,  AU M59 and AU M60 are related with Agreement and  Certainty.  Additionally, AU 4 alone can signal the state of concentration. AU 7 (Eyelid tightening) might be associated with concentration \cite{stratou2017refactoring}, and so in the context where people focus on a computer screen. AU14 (Dimpler) is found to be a facial indicator that periodically occurs over time and coincides with  thoughtful state \cite{littlewort2011automated}.  In the case of EMS, facial expressions also play important roles while  people are thinking, calculating, retrieving information from long term memory, and performing tasks that need meta-cognition  \cite{jackson2009enhanced}. \cite{dimberg2000unconscious}. 
Additionally,  open lip smile, closed lip smile, laughing loud are important communication features that have a big influence in the recognition of people’s mental states. \cite{lanzini2013different} and \cite{schneider1992preschoolers} suggest that different kinds of smiles and laughs are used mostly as for the detection of different EMS, such as delights, nervousness, confidence, interest, and engagement.



As we see that facial features are very important and significant for identifying the EMSs the people are going through, we  would need a robust framework  for automatic detection of EMSs from facial features. Automatic detection can help artificial tutors \cite{autotutor}, artificial training agents \cite{hoque2013mach} to understand users’ EMSs  while involved in the teaching environment or social interactions. A number of works are reported in last decade to estimate Epistemic Mental States (EMS) from facial features based on their underlying relation with particular EMSs. Particularly, there are couples of works that have performed automatic detection of learning related EMSs (i.e. confusion, engagement, interest, concentration,confusion, boredom). However, the most of them focus on on  particular Action Units (AU) while students are engaged in learning.  McDaniel et al. \cite{mcdaniel2007facial1} identifies the AUs that are strongly correlated with boredom, confusion, delight, frustration, and neutral. They showed that  AU4 (Brow Lowerer), AU7 (Lid Tightener),  AU12 (Lip Corner Puller), and AU 45( Eye Blink) are strongly correlated with boredom;  AU1 (Inner Brow Raiser), AU4, AU7, and AU12 are correlated with confusion. On the other hand,  AU4,  AU7, AU12, AU 25 (Lips Part), AU 26 (Jaw Drop), and AU 45 are responsible to interpret the emotion as delight. However, AU 12 and  AU 43 (Eye closure) indicate frustration. In later publications, SD Mello and his fellow researchers use FACET(commercial version of CERT \cite{littlewort2011computer} to detect above mentioned AUs while students were learning a programming language. Then they  perform AU modeling to detect the EMSs.

Additionally, in \cite{bosch2016detecting}, Nigel Bosch  et al. detect engagement, frustration, confusion, and boredom through the AUs while the AUs are identified by FACET. They are able to identify spontaneous judgement of confusion and frustration with kappa = 0.221 and 0.232, respectively. However, the results are not satisfactory because of low recall and low precision rate. The work \cite{bosch2014s} proposes supervised machine learning model where the authors build feature vectors from AUs and  their intensities. Then, they use these feature vectors in different Weka-based machine learning algorithms to predict the affective states. They build one detector for each states. So for confusion, they can discriminate confusion from other states. Same way, they build detectors for other states. They reported AUC measure for their performance: $Boredom = 0.610$, $Confusion = 0.649$, $Delight = 0.867$, $Engagement = 0.679$, and $Frustration = 0.631$ with supervised classification techniques. However, their data is very limited. Bousmalis et al. \cite{bousmalis2013towards} provided a detailed survey of the research works that analyze multimodal cues and nonverbal signals to automatically detect agreement and disagreement. They refer to a number of research papers that include facial action units from FACS , body posture, hand gesture, eye movements, and prosodic features. However, most of these research don’t  estimate the intensities of the Agreement/disagreement.  Mehu et al. \cite{mehu2014multimodal} have analyzed the audio visual recordings of debate sessions to extract nonverbal cues and audio visual signals that are strongly related Agreement/disagreement. However, this work doesn’t include any dyadic conversation. In another work \cite{poggi2010agreement}, the authors analyzed the agreement in  video fragments of political debates. Video fragments are used from Canal 9 \cite{vinciarelli2009canal9} and AMI corpora \cite{carletta2003nite}.  The researchers have used lingustic and verbal cues along with body postures (head movements and eye gaze) to detect “Agreement” in the video fragments of political debates.

Furthermore, \cite{pachman2016eye} proposes a machine learning model to facilitate early detection of confusion by tracking the eye movements. They collect eye-tracking data when the subjects are involved in solving puzzles in the computer screen. The authors try to find the correlation between eye fixation time  to the screen and problem solving time to identify how  subjects perceive confusion during the experiment. The paper \cite{grafsgaard2011modeling}  proposes an approach to identify EMSs from learners’ facial expressions when the learners are involved in a computer-mediated human task-oriented tutorial dialogues.  Specifically, they focus on particular Facial Action Units to detect the EMSs. The authors of \cite{krithika2016student} demonstrate a framework that can estimate the concentration level based on  the analysis of head-movements, eye gaze, and eye-lid movements etc. There are a number of research studies that have studied emotional states that students go through in learning environments. In the paper \cite{d2018cognitive}, the authors suggest  that 11 cognitive emotions: enthusiasm, interest, surprise, curiosity, concentration, attention, disappointment, boredom, perplexity, worried, frustration are perceived by the students when they are engaged in e-Learning activities. Histogram of Oriented Gradients (HOG) based features are extracted from the e-Learning video frames and  the features are then supplied to multiclass support vector machine (M-SVM). The M-SVM detects cognitive emotions in the frames. In the work \cite{de2019engaged}, the authors have analyzed the Engagement from  facial features, facial action units, and eye gazes during learning sessions. They have used facial analysis toolkit Openface 2.0 to extract the visual cues from the participants’ face. Then they have used Long Short Time Memory (LSTM)  to model the temporal evolution of the visual cues in order to estimate the Engagement intensities. However this study doesn’t include the analysis for dyadic context.  The work \cite{d2016emotions} describes how students experience positive and negative emotions  during the e-learning environment and how these emotions impact students’ engagement with the topics.

A number of researches also analyze EMSs during group/dyadic conversation. Emotion elicitation using dyadic interaction consists of dyads that are engaging in a series of unrehearsed and minimally structured  conversations in a controlled environment. This procedure helps elicit spontaneous emotions as well their temporal course of emotion. Example of dyadic contexts are: romantic interaction, sibling interaction, and patients/therapist interaction. In romantic  interaction, couples would engage in dyadic conversation where they discuss their feelings, like, dislikes, and expectations etc. In sibling interaction two siblings talk about their expectations. In patients/therapist interaction, they talk about their disease, medications, health problems. The dyads go through different kinds of emotions such as  feeling positive, romance, happiness, agreement, interest in romantic context. In sibling context, dyads would feel  jealousness, certainty, thoughtfulness etc. In the health service context, emotions like care, worry, frustration, certainty might be present. The emotions evolve over time as the interaction continues. Hayley et al. proposed an automated framework to estimate cohesion of a small group using the visual cues from participants \cite{hung2010estimating} in social conversation. The authors analyze the body postures, hand gestures along with speech to detect cohesion.   Sometimes, head nods indicate the interest and concentration about some knowledge-based discussion. As part of the back channeling context, the user's head nods are not only used to express an agreement, but also to display interest or enhance communicative attention.  However, detecting head nods in natural interactions is a challenging task as head nods can be subtle, both in amplitude and duration. In the study \cite{nguyen2012using}, they propose to use the dynamics of head gestures conditioned on the person's speaking status.

The evolution of  epistemic mental states is sometimes determined by the conceptual capabilities of the dyad subject and by his/her context, i.e., the set of knowledge that the subject possesses and is able to apply, in addition to the properties of the cognitive task the subject is confronting \cite{arango2014nature}. These factors (subject’s conceptual capacity, context, and the properties of the cognitive task) determine the content, intensities, and characteristics of  epistemic feelings. Moreover, the evolution of EMSs over time also depends on dyad’s prior knowledge \cite{michaelian2014epistemic}. For example, if a dyad subject is asked about some knowledge question (i.e: What is the capital of Italy?), the subject might  have a feeling that he/she is in possession of the information but is unable to access or recall it fully from memory: he/she would experience the state of uncertainty. However, even if he/she is unable to recall the exact answer, the subject has access to partial information such as the first letter, or what the city name sounds like. The subject gradually gains his/her certainty level high enough through his/her metacognition. And finally, he/she is able to discriminate the correct answer among different possibilities with highest certainty.


Though the most of the emotional state recognition research  has been performed on nonverbal communication, there are some research works that propose automatic frameworks to detect epistemic states from verbal communication as well as nonverbal ones. A number of research works have been performed where researchers analyze the  emotion-specific language pattern to predict the epistemic states. The authors of the work  \cite{inproceedingsLinguistic} aim to analyze  appropriate verbal labels  of real-life dialogues to predict  the presence of different emotions. They analyzed the transcripts of thousands of client interactions with call center employees. Riley et al. \cite{forbes2004predicting} examined  the efficacy of different types linguistic features for automatically estimating  student emotions in human-human spoken tutoring dialogues.  They have used dialogue turns and linguistic features to classify student emotions by using Adaboost classifiers. In \cite{lee2002combining},  the authors have  adopted  an information-theory based algorithm to identify  emotion salient keywords in the transcripts of  dyadic interaction. Then they have combined these keyword features towards acoustic features to classify the emotions that are present in the interaction. \cite{borras2011perceiving} analyze an important research question whether the perception of  Certainty/Uncertainty are characterized by verbal features, prosodic features, or facial features using Linear Mixed Model. They found that  all three cues have a significant impact on perception of Certainty. However, the visual cues: (facial gestures) have more   dependence on Certainty prediction than linguistic as well as prosodic features.
Finally, the work \cite{contVA} shows that continuous Valence and Arousal can be predicted reliably from facial behavior. They have used Bidirectional Long Short-Term Memory neural networks to predict Valence and Arousal from spontaneous Sensitive Artificial Listener (SAL) dataset. In every face-to-face meeting - even if participants are visually impaired - an order of dominance is established after a short period of time \cite{lambDom}, \cite{dominanceEyeGaze}. In the context of dyadic interactions, visual indicators of dominance are observed in the face through factors like facial expressions, gaze, and head pose \cite{dominanceEyeGaze}. Hence, dominance is expected to be predicted from visual cues of the partner’s face videos. However, the works \cite{contVA} do not include robust tracking of the visual features without any human intervention. They correct the face and shoulder tracking manually when the tracking fails. Such a system is quite unreliable for automatic emotion prediction in a natural environment. The desired system should automatically correct the facial tracking when it fails and should be able to make the prediction in real-time. 

Therefore, even if a lot of studies examine emotions and affective states, it is very hard to find research that investigates the complex relations between EMSs and facial features. Based on the research works described above, it is easy to note that the prototypes suffer from several kinds of limitations due to a lack of understanding of the relations between EMSs and facial features. 

\section{Facial Features and  Epistemic Mental states} \label{sec:overview}

In this section, detail methods are proposed to answer the research questions indicated in the introduction. The proposed methods and the analysis results are presented using the all videos of the SEMAINE dataset \cite{semain}. 

\subsection{Facial Features} \label{sec:Facial Features}
Facial features that define the facial behavior efficiently are extracted from the face and used in the analysis. Moreover, head pose is inherently related to person's mental states. Hence, the following facial features and head pose were extracted: 
\begin{itemize}[noitemsep,nolistsep]
\item Height of Inner Eyebrow (inBrL, inBrR),
\item Height of Outer Eyebrow (otBrL, otBrR),
\item Eye Opening (EyeOL, eyeOR),
\item Height of Outer Lip Boundary (oLipH),
\item Height of Inner Lip Boundary (iLipH),
\item Distance Between Lip Corners (LpCDt),
\item Head Pose (Yaw, Pitch, and Roll)
\end{itemize}

\begin{figure*}[!h]
\centering
\includegraphics[]{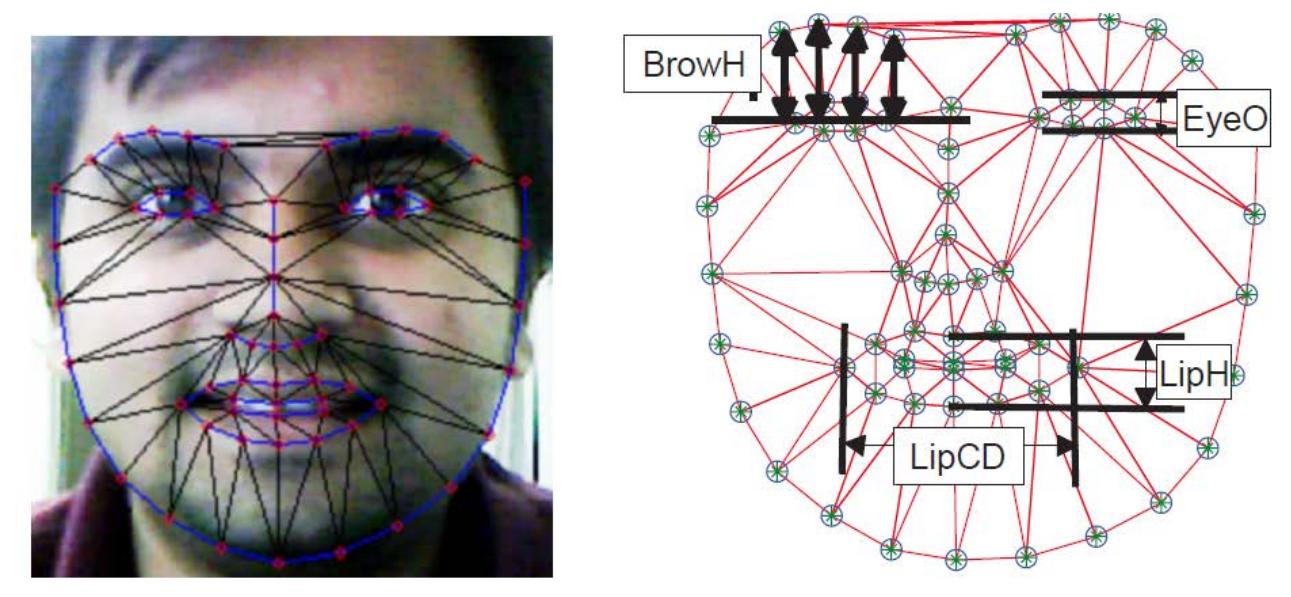} 
\caption{Face Tracker and Feature Extraction from face}
\label{FeatureExtraction}
\end{figure*}

Facial features and pose information (Yaw, Pitch, and Roll) are extracted using a Constrained Local Model (CLM) based face tracker that uses the algorithm of Saragih et al. \cite{Saragih3}. It can track 66 landmark points in each frame containing a face. The algorithm works by fitting a parameterized 3D shape model of face with a target image. The distance based features (height of inner eyebrow, height of outer eyebrow, eye opening, height of outer lip boundary, height of inner lip boundary, and distance between lip corners) have been extracted from the tracked landmarks points as shown in Fig \ref{FeatureExtraction}. To do so, a canonical reference shape was used which is actually the mean of the shapes in different facial expressions. The features were extracted as a ratio of the distances between appropriate landmarks in the reference shape and the tracked shape after removing the global transformations. The head pose information has been also    estimated from the 3D-tracked shape as the tracked shape contains the 3D rotation information.

\begin{figure*}[!t]
\centering
\includegraphics[width = 6 in]{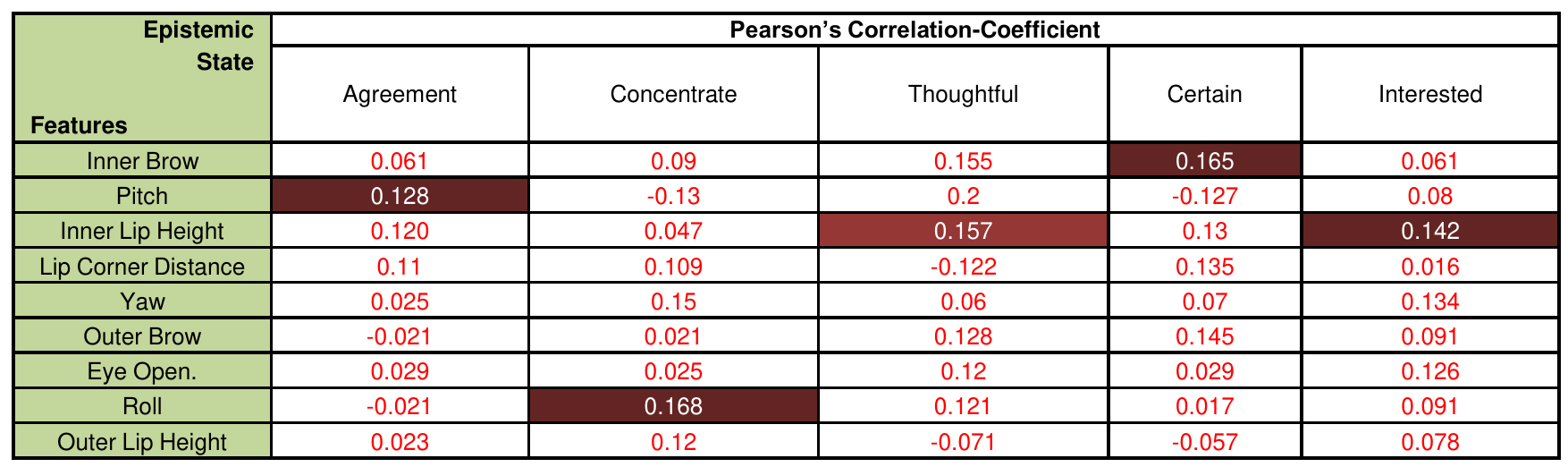}
\caption{Pearson's correlation-coefficients: Linear Relations between Features and Epistemic States}
\label{LinearRelations2}
\end{figure*}

\subsection{Complex Interplay between Features and Epistemic States}
\label{sec:ComplexInterplay}
Unlike the basic expressions, people exhibit subtle and critical changes in their faces during the EMSs. These subtle changes make the prediction of the epistemic states quite challenging. Therefore, it is utmost necessity to understand the interplay between facial features and the EMSs in order to model the emotion prediction framework. Additionally, quantifying the relations will help to identify the mostly related facial features to the particular states. The research \cite{atkinson2005visual} suggests that the complex states are recognized by the sighted people disproportionately by processing the visual cues from the eye and mouth region of the face. For instance, social communications with eye to eye contact provide information about concentration, confidence, and engagement. Smiles are universally recognized as signs of pleasure and welcome \cite{reviewJournal}. In contrast, looking away for a long time is perceived as lack of concentration or break of engagement. Frequent yawning is an indicative symbol for boredom. Yaw (left/right) and pitch (up/down) have stronger relations with agreement. In contrast, lip corner distance and pitch are more important to analyze certain feelings.




Following two subsections, we present our methods to answer our first research question: 

	\textbf{$\bigstar$}	How facial features relate to epistemic mental states? What are the characteristics of these relations?
	
	\subsection{Linear Relations} \label{LinearRelations}
We start the analysis of the interplay  by quantifying the linear relations between features and mental states.  To find the linear characteristics of the facial feature vs epistemic states relations, pairwise Pearson's correlation-coefficient are calculated and listed in figure \ref{LinearRelations2}.

Facial features are complex and dynamic. Due to the complex and convoluted characteristics of facial features and their temporal dynamics, the Pearson's correlation-coefficients are found  insignificant  to capture the complex relations between them. Hence, the results in figure \ref{LinearRelations2} show that the facial features have very small correlation-coefficients with corresponding states. The shaded cells show the highest coefficients for corresponding features and epistemic states. Accordingly, it is quite impossible to model the epistemic states using the linear relations with facial features. 

\subsection{Non-linear Relations} \label{subsec:NonlinearRelations}
Since, it is observed that the Pearson's correlation-coefficients are not significant enough to capture the interplay between facial features and states, we move forward to capture the non-linear characteristics of their associations to quantify the complex and temporal dynamics relations.

\begin{figure*}[!t]
\centering
\includegraphics[width = 6 in]{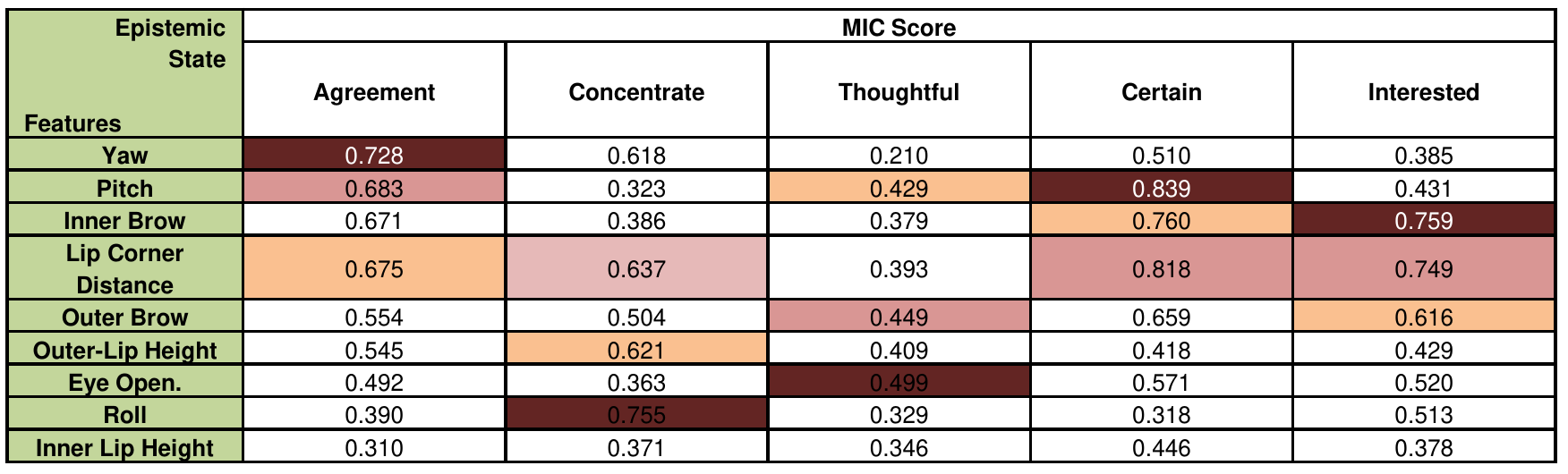}
\caption{Non-linear relations (MIC Scores)between  Features and Epistemic States}
\label{Nonlinearrelation}
\end{figure*}

\subsubsection{Ways to extract non-linearity}
\label{subsubsec:Waystoextractnon-linearity}
 To capture the non-linearity, the state-of-the-art data  analysis tool  "Maximal Information-based Non-parametric Exploration" (MINE) \cite{reshef2011detecting} developed in MIT and Harvard is used calculate the nonlinear coefficients. In particular, Maximal Information Coefficient (MIC) score from MINE quantifies the non-linear relations. MIC is a part of a larger family of MINE statistics, which can be used to identify and characterize non-linear relationships in large data \cite{reshef2011detecting}.  Before going to the analysis part, we would like to give a brief description about MINE and MIC in following paragraphs.
\subsubsection{What is  MINE and MIC?}
Finding and quantifying significant relationships between pairs of variables in large data sets is challenging and important \cite{reshef2011detecting}. The paper \cite{reshef2011detecting} presents a measure of dependence for two-variable relationships: the maximal information coefficient (MIC). MIC captures a wide range of relations: both functional and non-functional.

For an emotion data set with the number of facial features and  head pose   may contain important, undiscovered relationships. If  we do not know what kinds of relationships to analyze for,  it is quite impossible to  identify the important ones efficiently.  Data sets of this size are increasingly common in fields as varied as facial expressions, physiological signals, genomics, and bio-informatics. One way to begin exploring a large data set is to search for pairs of variables that are closely associated. To do this, MINE presents a measure (MIC) of dependence for each pair.  Then we can rank the pairs by their MIC scores, and examine the top-scoring pairs. The maximal information coefficient ( MIC) is based on the idea that if a relationship exists between two variables, then a grid can be drawn on the scatter-plot of the two variables that partitions the data to encapsulate that relationship.

\subsubsection{Facial Features, epistemic mental states, and MIC scores}
MINE software package is exploited to calculate the pairwise MIC scores. A matlab program is developed to calculate pair-wise MIC  for each feature with the epistemic states. The pairwise MIC scores are shown in figure \ref{Nonlinearrelation} where the left column indicates the facial features. From the figure \ref{Nonlinearrelation}, it can be easily noted that every feature has some non-linear relations with particular states.

In the figure \ref{Nonlinearrelation}, dark colored cells represent highest MIC score for particular states. Lighter colors represent second and third highest MIC scores.  Specific features are strongly related to particular states. For example, yaw (left/right) and pitch (up/down) have stronger relations with agreement state (0.728 and 0.683, respectively). Moreover, inner brow distance and lip corner distance have second best scores (0.675 and 0.671) with agreement. For the concentration: Rolling (0.551), lip corner distance (0.437), and outer lip height (0.412) have higher relations with concentration. Particularly, people roll their head to look around when they are not concentrating to their interlocutor. Since, the negative concentration-rating also provides information about the lack of concentration, head rolling has much more impact on concentration. Lip corner distance and outer brow raise are related to yawning that can also indicate the lack of concentration. 

Therefore, we propose to use the MIC scores (for corresponding EMS) as the feature weights when the facial features are applied to machine learning models. That means: MIC score-weighted facial features are fed to the models for the experiments.

\subsection{Temporal feature vs Epistemic States}
\label{sec:Temporalfeature}
Since epistemic states are discrete states and the states depend on the temporal pattern of the facial features, their temporal dynamics should be included in our analysis. In this section, we  perform necessary analysis to answer the following question:

\textbf{$\bigstar$}   What are the temporal features and their  temporal dynamics related to epistemic state of mind?  What levels of granularity do they interact that can be used for automated recognition of them in real-time?

\begin{figure*}[!t]
\centering
\includegraphics[]{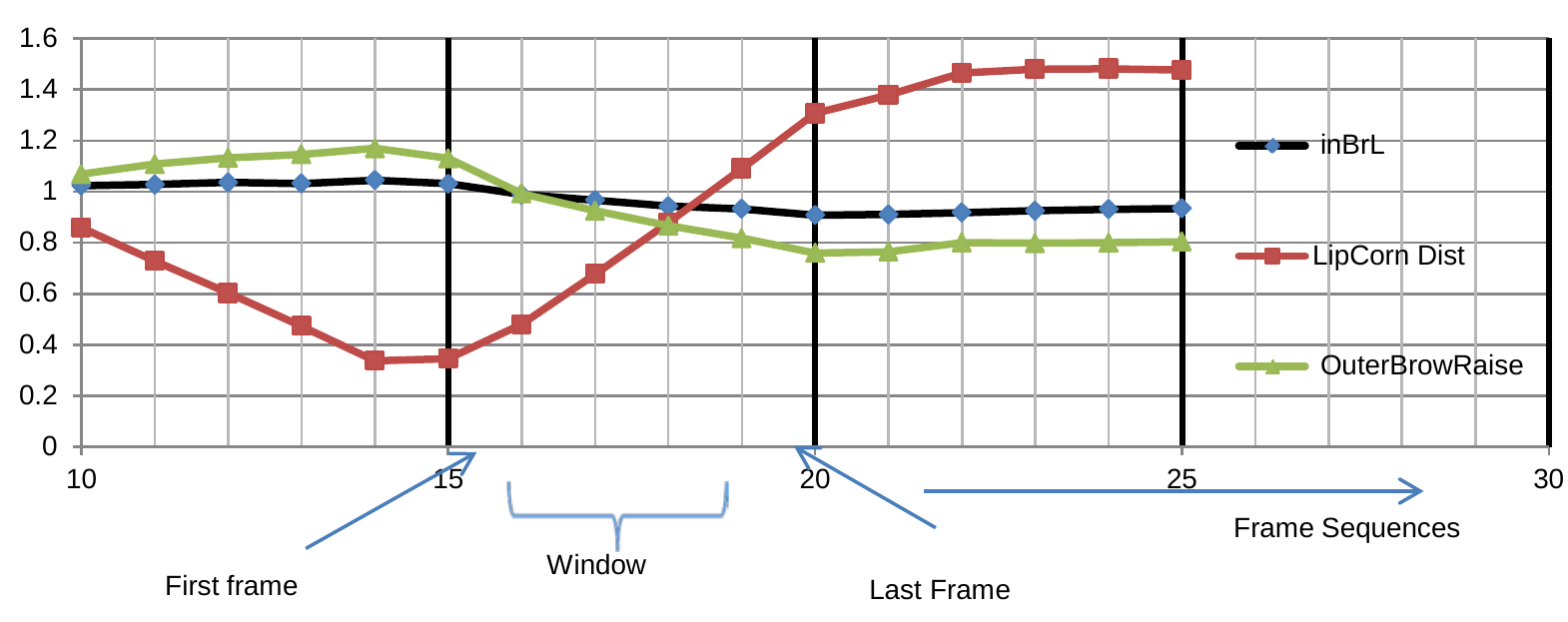}
\caption{Sliding Window technique with five frames to calculate temporal features}
\label{SlidingWindow}
\end{figure*}

To perform the analysis, we  use the simpler technique to provide real-time processing. We follow the sliding-window paradigm to compute temporal dynamics of the features. In this paradigm, the window is slided right one at a time to include next frame into the window and to discard the oldest frame of the window (figure \ref{SlidingWindow}). Using the sliding window paradigm, we explore and calculate a number of temporal features to capture the temporal evolution of the  features.  

This windowing technique has following advantages over others:
\begin{itemize}
\item Continuous analysis of a video or camera feed.
\item Fast computation.
\item Flexibility of changing window size to model spontaneous feature evolution.
\end{itemize}

\subsubsection{First order temporal features}
\label{subsec:Firstorder}
The first order temporal features (velocity, velocity directions) are calculated in a fixed window. The velocity of a particular feature within a window was calculated using the formula:

\begin{equation}
 V_{last} = \frac{F_{last}- F_{first}}{T_{last}-T_{first}}
\end{equation}

where $V_{last}$ is the velocity of the corresponding feature at the last frame. The sign of the $V_{last}$ signifies the rise or fall of the feature. $F_{last}$  and $F_{first}$  are the feature values in the last frame and the first frame of the window, respectively. $T_{last}$ and $T_{first}$ are the time of the last frame and the first frame. Therefore, the magnitude $V_{last}$ and its sign can provide the temporal information about that corresponding feature in that window.

After calculating the temporal features, we use MINE to calculate the MIC scores between computed velocity features and epistemic states. Unfortunately, for window size six, we have not obtained promising MIC score for the temporal features. Following figure shows the MIC scores between temporal features and the concentration.

\begin{figure}[!h]
\centering
\includegraphics[width = 4 in ]{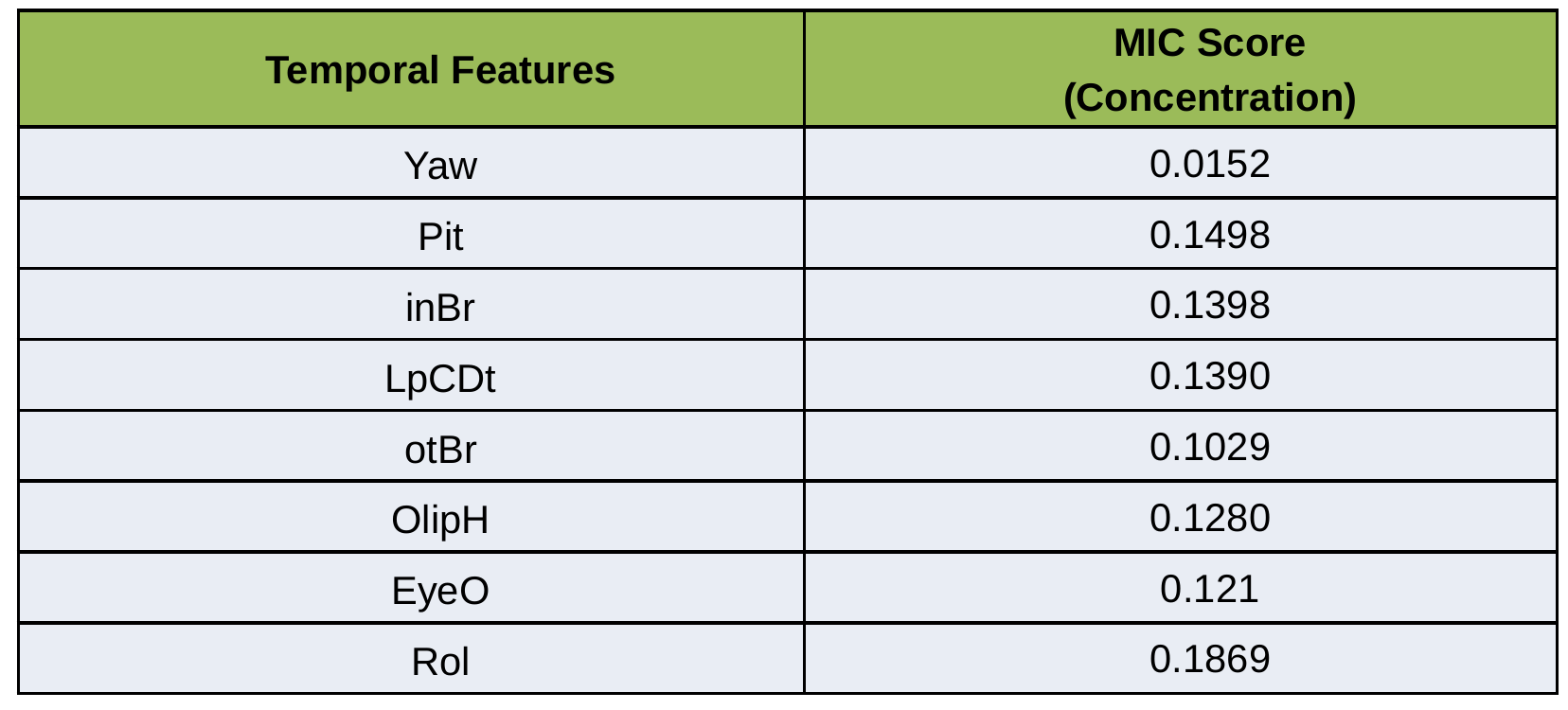}
\caption{MIC Scores between Temporal Features and Concentration}
\label{MICScoresTemporal Features}
\end{figure}

\begin{figure}[!h]
\centering
\includegraphics[width = 4.5 in]{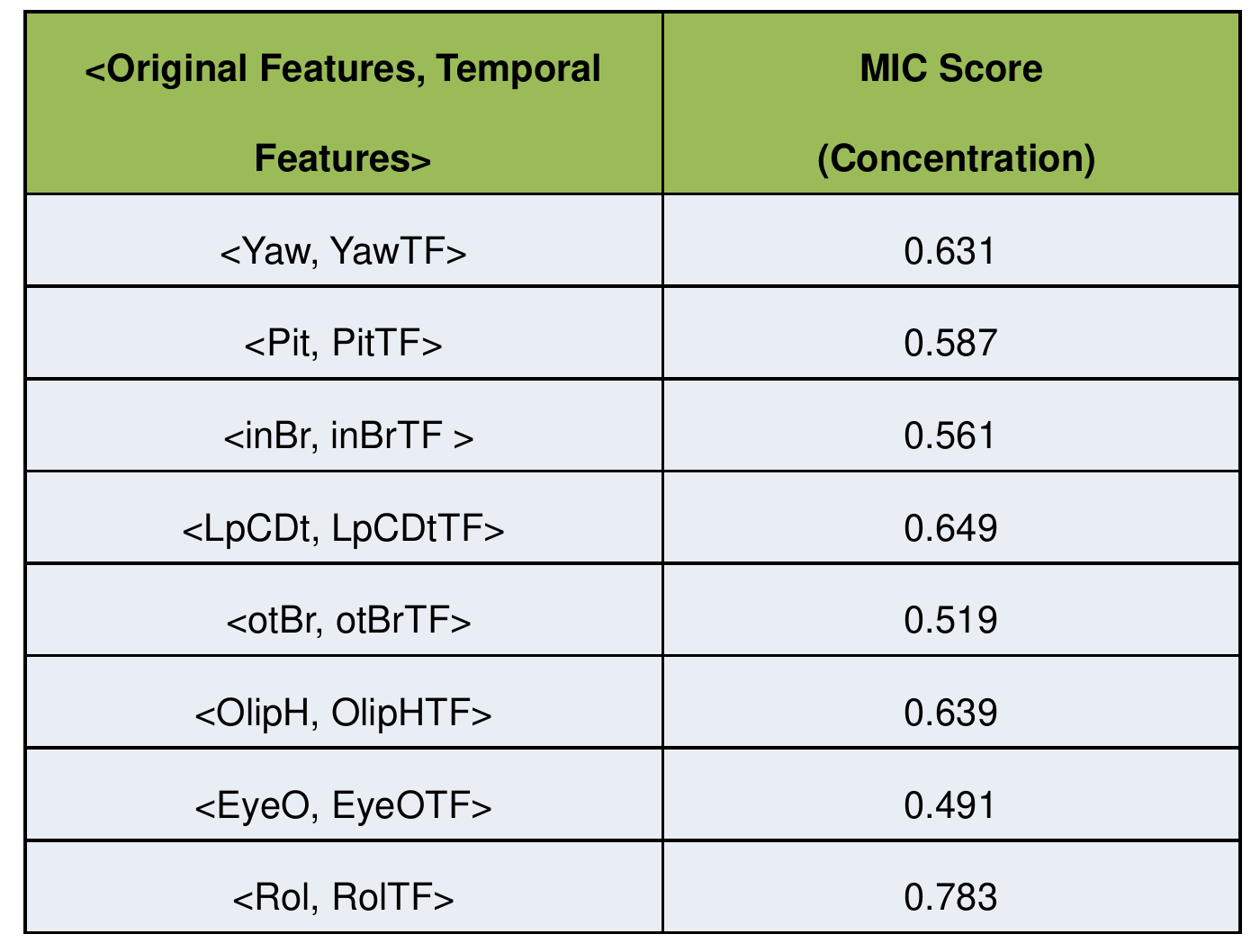}
\caption{MIC Scores for tuples ($<Original Features, Temporal Features>$) against Concentration}
\label{touples_Concentration}
\end{figure}

To confirm the results, number simulations were performed with various window sizes (10, 20, 30, and 40). However, the MIC scores did not improve at all. The lower scores pushed us to investigate why temporal features have lower MIC scores. To do that, we created a tuple $<feature, velocity>$ that combines the facial features and corresponding velocity. Then, we have calculated the MIC again with the tuples and the mental states. Interestingly, the tuples have very high MIC scores (figure \ref{touples_Concentration}) with the states and we have found the answer.  

The MIC scores shown in the figure \ref{touples_Concentration} verified that temporal features are important while used with original features. Therefore, we also use these MIC score-weighted temporal facial features  while feeding them to the models for the experiments.


\section{Intensities of Epistemic Mental States}
\label{sec:intensities}
The epistemic mental states are evolving over time based on the social context \cite{arango2014nature} \cite{michaelian2014epistemic} \cite{da2009epistemic}. Consequently, the dynamics of the facial features  also change during the evolution. However, the relation between dynamics of facial behavior and evolution of mental states are not studied in the recent literature. The analysis done in the previous section could not provide these clues. Following subsection describes the different regions of the  the mental state evolution.

\subsection{Regions  of Epistemic Mental States}
\label{subsec:regions} 
The EMSs evolve over time based on the social context. Concentration can increase in successive conversation while the topics of becomes more important.  The intensities can be also sustained for some time period or can be decreasing. 

\begin{figure*}[!t]
\centering
\includegraphics[]{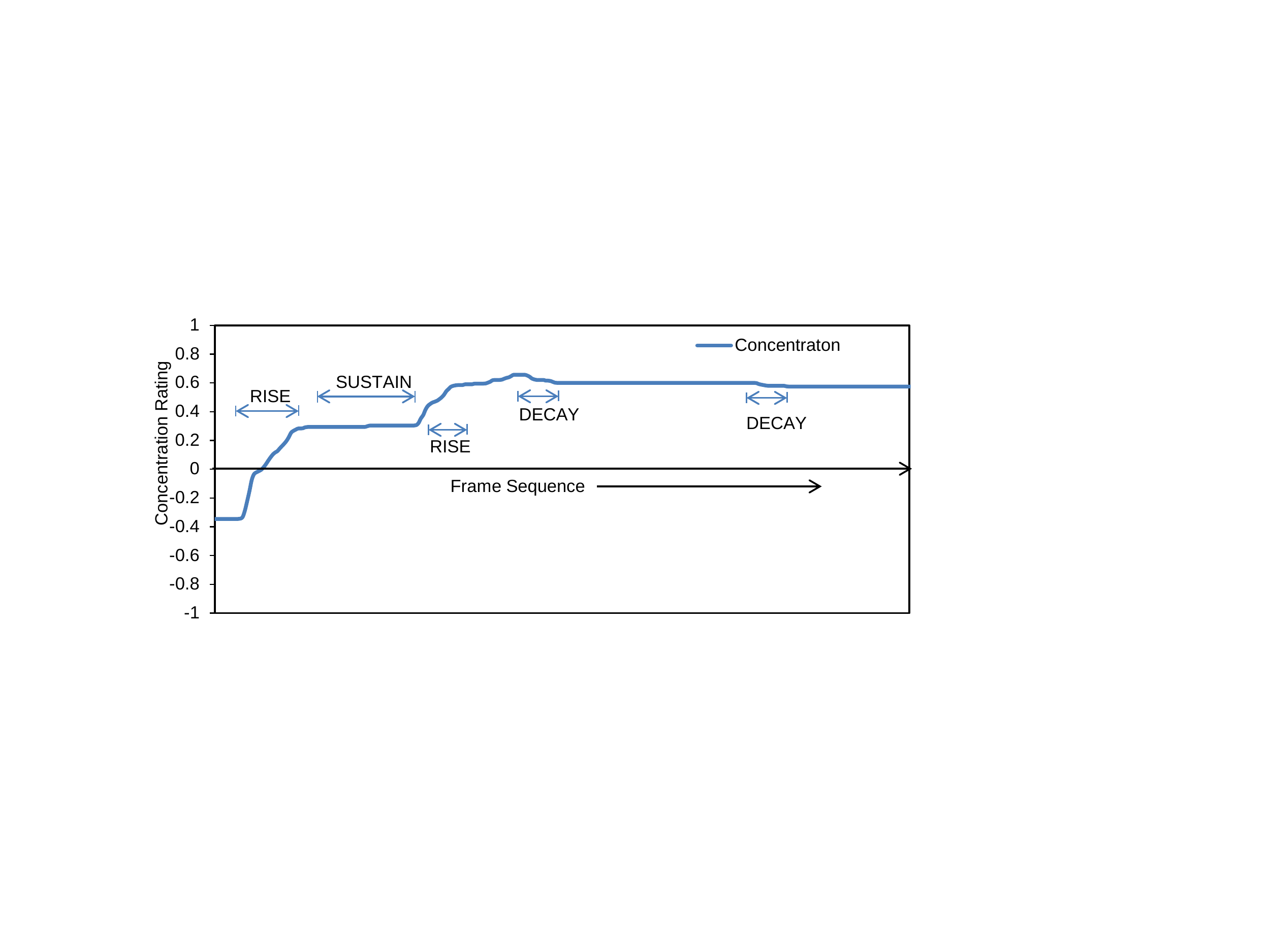}
\caption{Different regions of Epistemic States in continuous frame sequence}
\label{Differentregions}
\end{figure*}

Figure \ref{Differentregions} shows the different regions of concentration in a dyadic interaction video from SEMAINE. 

During the RISE, the concentration is continued to rise, then it sustains for some time followed by another rising.  Then it decays to some extent before being sustained for a longer period during important discussions with the conversation partner. Henceforth, the features and their temporal evolution may not be modeled equally for whole video sequence. Specifically, we need to get the answers of following research questions:

	$\bigstar$ Do the temporal features have the same impact on concentration in the whole sequence?
	
	$\bigstar$ Do RISE, SUSTAIN, and DECAY regions have different relations with temporal features? 

\subsection{How regions are related to temporal features?}
\label{subsec:Automated}

To answer the questions, several additional features are derived from the original facial features to capture the temporal dynamics. We have calculated three types of derived features from the facial features from SEMAINE videos of concentration. We follow the steps listed below:
\begin{enumerate}
\item	Calculate following features:\\
  	\begin{enumerate} [label=\roman*., itemsep=0pt, topsep=0pt]
  		\item Original features (weighted by their corresponding MIC scores)
		\item Velocity of  features (weighted by their corresponding MIC scores)
		\item	Up, down, and unchanged (+1,-1,0) events
		\item	Events multiplied by velocity 
		\end{enumerate}
\item	The regions are labeled with ground truth  (RISE, SUSTAIN, DECAY)
\item	Predict concentration for each of the region using different support vector regressions.
\end{enumerate}
Then the rising, steady, and falling regions are manually annotated to generate the ground truth. The regions are labeled with corresponding regions (RISE, SUSTAIN, FALL, respectively). The features of corresponding regions are concatenated with the labels to form Weka file to predict the concentration rating in the corresponding region using support vector regression (SVR). After estimating the concentration rating, we calculate correlation coefficient (CoERR) between ground-truth  rating and predicted  rating for concentration. We use the CoERR as the performance metric throughout the paper.

\begin{figure}[!h]
\centering
\includegraphics[]{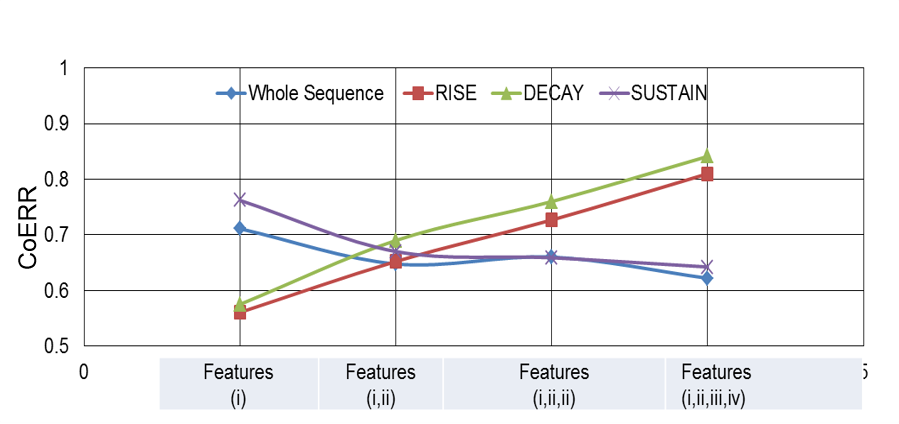}
\caption{Prediction of Concentration intensity in different regions with different features}
\label{PredictionofConcentration}
\end{figure}

The number shown in figure \ref{PredictionofConcentration} illustrates that the use of original features along with first-order temporal features is not efficient enough to model concentration in a whole video. Specifically, they are very useful while used in emotion changing regions (RISE or FALL). On the other hand, temporal features has diminishing impact in SUSTAIN region. instead, original features are found robust in the SUSTAIN region.

Then, we use increasing number of types of features to incorporate more temporal information into the feature vectors. The figure \ref{PredictionofConcentration} shows the impact of the temporal features on different regions of the concentration. As we go from left to the right of the X-axis, more temporal features are included to the model for the corresponding region.  The plot suggests that temporal features are very useful while used in the RISE or DECAY region. Intuitively, as the intensity of a particular epistemic state changes, the facial behavior evolves accordingly. For example, the head may be the tilted or outer lip height would be in the normal position from yawning. Additionally, original features alone are optimal for SUSTAIN region. Addition of temporal ones have diminishing impact on concentration during the SUSTAIN region. It is easy to note  that the performance for SUSTAIN region from 0.76 to 0.65 while we use feature sets i,ii,iii, and iv. 

\begin{figure}[!h]
\centering
\includegraphics[ ]{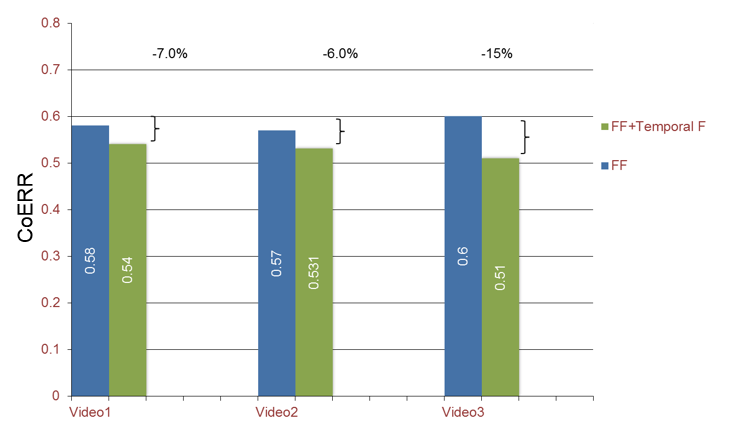}
\caption{Performance decremented for Concentration prediction without any region information (10-fold cross validation on three concentration videos)}
\label{pictureVideoWORegion_Coerr}
\end{figure}

The same observation has been found in our pilot test. When we try the support vector regression (SVR) model  with all features (i,ii,iii,iv) on three concentration video episodes, prediction performance (CoERR)  decreased in figure \ref{pictureVideoWORegion_Coerr} due to the lack of region information. SVR gets confused with no information of SUSTAIN, RISE, and DECAY regions.

\subsection{Automated Modeling of  Epistemic States}
\label{sec:AutomatedModeling}
The results of analysis performed above motivate us to step forward towards developing an automated framework to predict the intensity of an epistemic state from facial features and their temporal dynamics.  Since the facial features have different relations with the epistemic state in various regions; we are interested in recognizing the region first.  

Then, the support vector regression (SVR) \cite{vapnik1997support} associated with that region is proposed to be applied to predict the intensity of the epistemic state. To model regions, the all facial features, temporal features, and derived features are calculated and concatenated to build the feature vectors. So, we need to answer following research question:

\textbf{$\bigstar$} How can we detect the different regions of concentration from features?

\begin{figure}[!h]
\centering
\includegraphics[width = 6 in ]{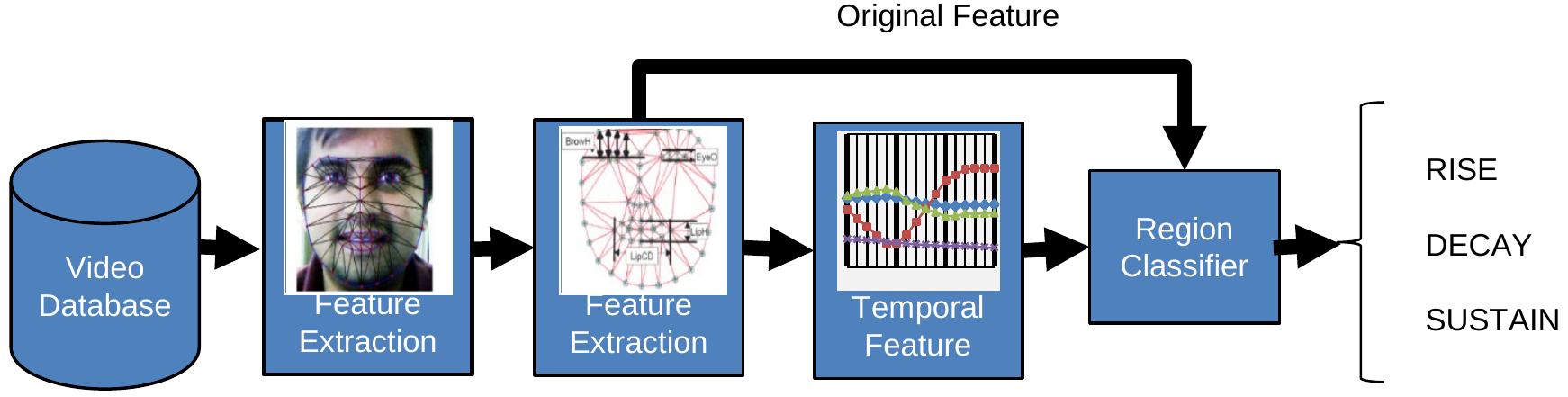}
\caption{Framework for Region Classification from features and their temporal dynamics}
\label{RegionClassificationFramework}
\end{figure}

The results found in the above subsection motivate us to build classifier models that can recognize the RISE, SUSTAIN, and FALL regions automatically from the original and derived features. The simulations have following input and output.
\begin{itemize}[noitemsep,nolistsep]
\item	Input:  Original and temporal features ( weighted by their corresponding MIC scores), derived features
\item	Output: the region (RISE, DECAY,  or SUSTAIN)
\end{itemize}
The figure \ref{RegionClassificationFramework} shows the schema diagram for region classification. Tree-based classifier was used to classify the regions. 

\begin{figure}[!h]
\centering
\includegraphics[width = 5 in ]{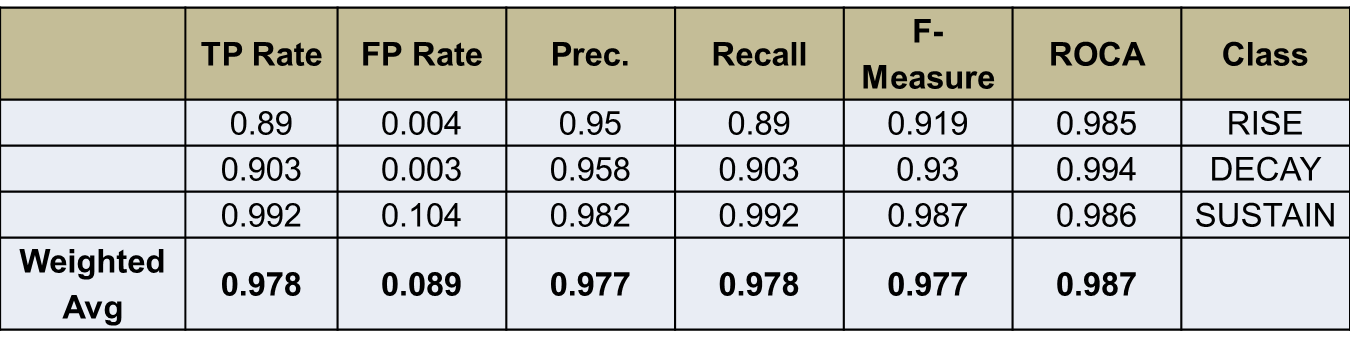}
\caption{Performance of Random Forest classifier for region with 10-Fold Cross Validation}
\label{RandomForestclassifier}
\end{figure}

The figure \ref{RandomForestclassifier}  enlists the performance for region classification. It can be easily noticed that the average ROC Area (Area under Receiver Operating Characteristic curve) is reasonable enough for robust region classification. Hence, if we can recognize the region effectively, the region information would facilitate to select out appropriate feature set/feature combination among original features, temporal , and derived ones  for concentration prediction. 

\begin{figure*}[!t]
\centering
\includegraphics[width = 6 in]{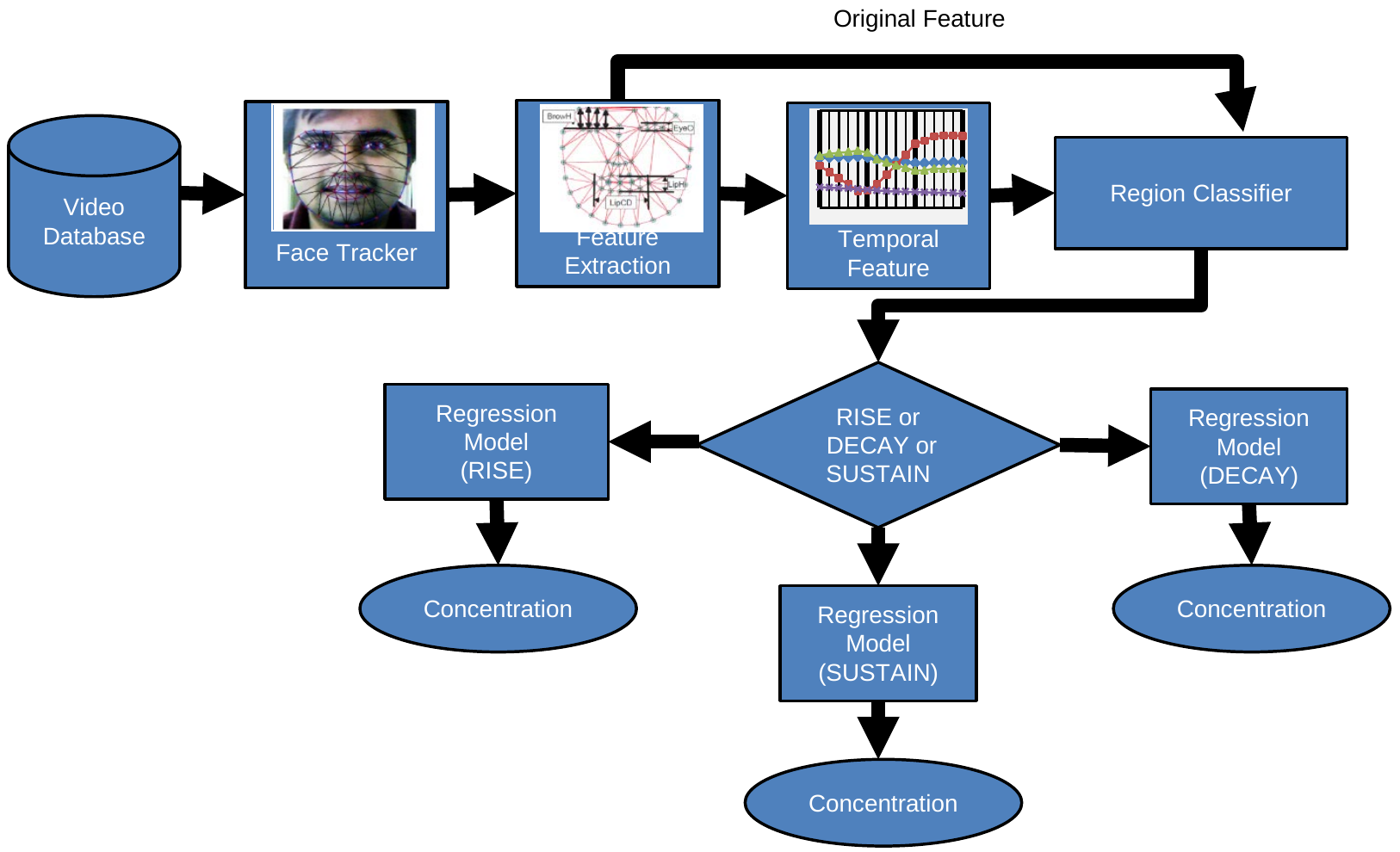}
\caption{Concentration Prediction Model that includes region classifier as an intermediate step}
\label{concentrationPredictionModel}
\end{figure*}

\begin{figure}[!h]
\centering
\includegraphics[width = 4 in ]{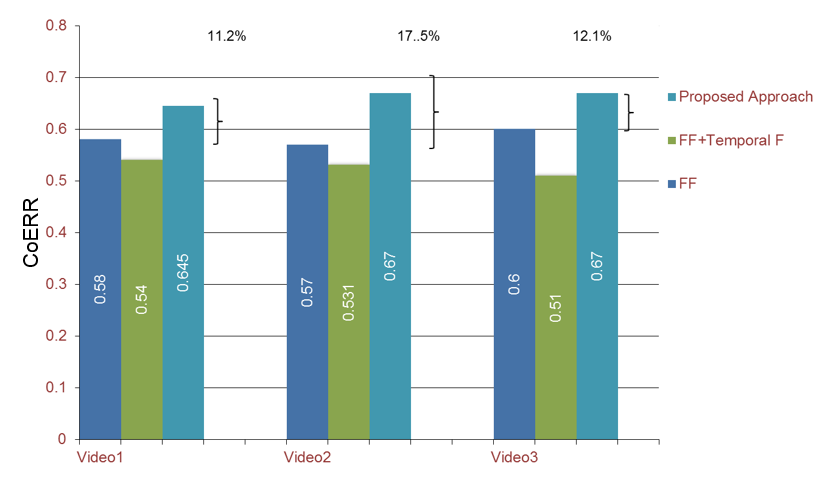}
\caption{Performance improvement for Concentration prediction with the region classifier -  (10-fold cross validation on three concentration videos)}
\label{pictureVideoWithRegion_Coerr}
\end{figure}

After recognizing the region, corresponding support vector regression model with appropriate feature combinations are designed to predict the concentration intensity. After using the model shown in \ref{concentrationPredictionModel}, we get the improved prediction performance in the three concentration video episodes described above. From figure \ref{pictureVideoWithRegion_Coerr}, it is easy to notice that the COERRs are significantly improved by $11.2\%$, $17.5\%$, and $12.1\%$ for video1,video2, and video3, respectively.


\section{Episodes of Epistemic states}
\label{sec:Episodes}

\subsection{Duration of the Epistemic Mental States}
\label{subsec:DurationEpistemicStates}
As discussed in the introduction section, epistemic mental states are lengthy than the categorical ones. Moreover, the duration of the states are not same for all of the states. 
\subsection{Window Modeling}
\label{sec:windowmodelingn} 
In order to analyze the differences between various epistemic states in terms of duration, we did vary the window size accordingly from 5 to 100. Please see the result section for graphs and description in detail.

\subsection{Emotion vs. Emotion Relations}
\label{sec:EmotionvsEmotion}
\begin{figure}[!h]
\centering
\includegraphics[width = 3.5 in]{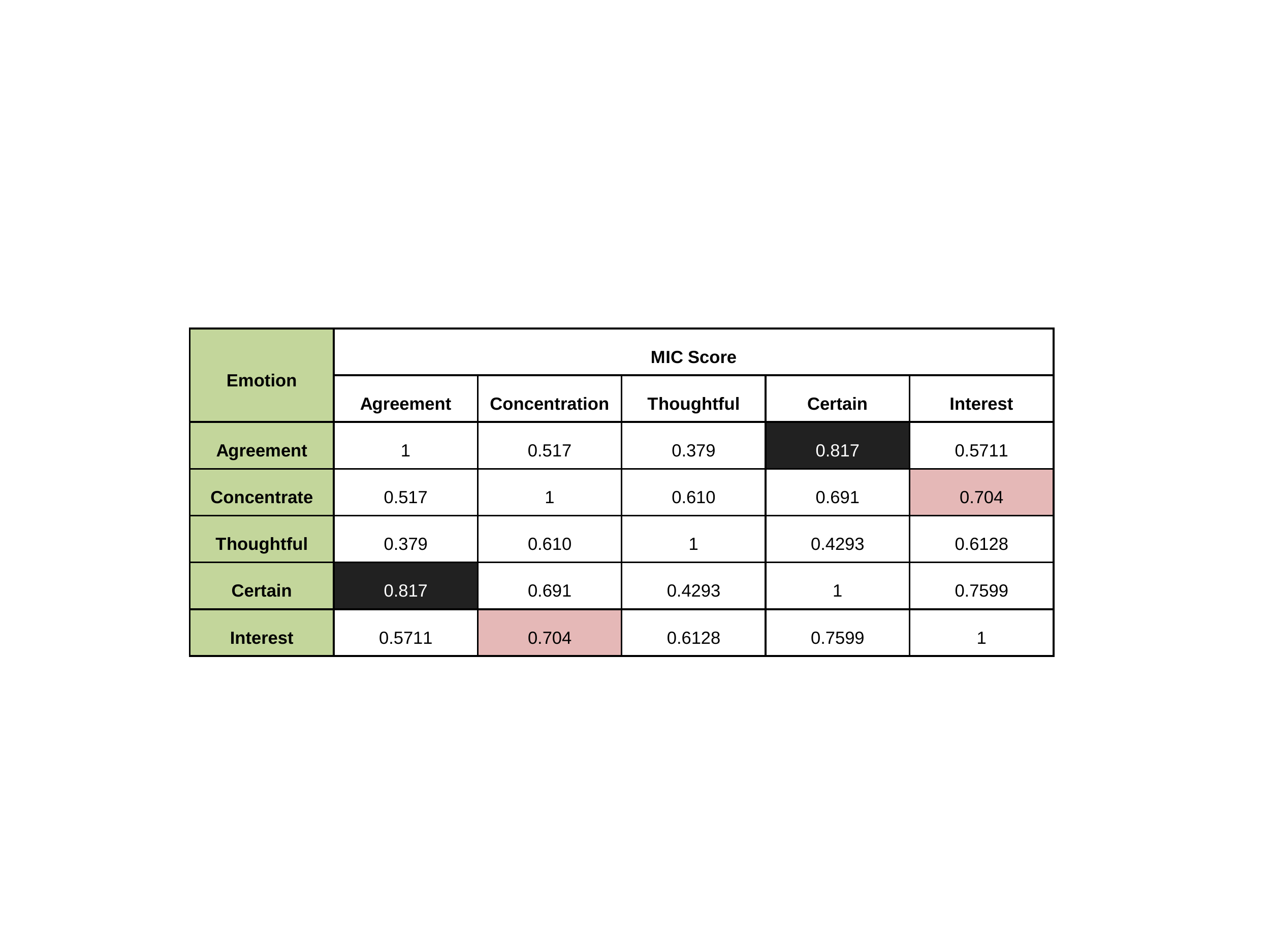}
\caption{MIC Scores between Emotions}
\label{MICScores(EmotionEmotion)}
\end{figure}

Aside from the relations between features and epistemic states, the mental states are also related to each other due to the dyadic context. Intuitively, while the dyads are becoming interested in some topics, they concentrate to the conversation. Moreover, agreement and certain are strongly related to each other due to the certainty about their decisions. To quantify the non-linear relations, MIC scores are also computed between the epistemic states and they are illustrated in figure \ref{MICScores(EmotionEmotion)}.	


\section{Databases used in the Performance Evaluation}
\label{subsec:database}

We use the SEMAINE videos \cite{semain} that contains spontaneous data capturing the audiovisual interaction between a human and an operator undertaking the role of an avatar with four personalities: Poppy (happy), Obadiah (gloomy), Spike (angry) and Prudence (pragmatic). The audiovisual sequences have been recorded at a video rate of 25 fps (352 x 288 pixels). The SEMAINE consists of audiovisual interaction between a human and an operator undertaking the role of an agent (Sensitive Artificial Agent). SEMAINE video clips have been annotated with couples of epistemic states such as agreement, interested, certain, concentration, and thoughtful with continuous rating (within the range [1,-1]) where -1 indicates most negative rating (i.e: No concentration at all) and +1 defines the highest (Most concentration). Twenty-four recording sessions are used in the Solid SAL scenario. Recordings are made of both the user and the operator, and there are usually four character interactions in each recording session, providing a total of 95 character interactions and 190 video clips.   



A key objective of the Solid SAL scenario is to record behaviors (mainly nonverbal) that a human operator shows in fluent face-to-face conversation, including their relationships to user behavior-notably back-channeling, eye contact, various synchronises, and so on. Users are encouraged to interact with the characters as spontaneously as possible. The audiovisual sequences have been recorded at a video rate of 25 fps (352 x 288 pixels) and at an audio rate of 16 kHz. The recordings are made in a lab setting, using one camera, a uniform background and constant lighting conditions. The SAL data has been annotated by a set of coders who provided continuous annotations with respect to the above epistemic mental states using the FeelTrace annotation tool \cite{tomkins1962affect}. Feeltrace allows coders to watch the audiovisual recordings confined to [-1, +1], to rate their impression about the emotional state of the subject. 

\begin{figure}[!h]
\centering
\includegraphics[ width = 3.5 in]{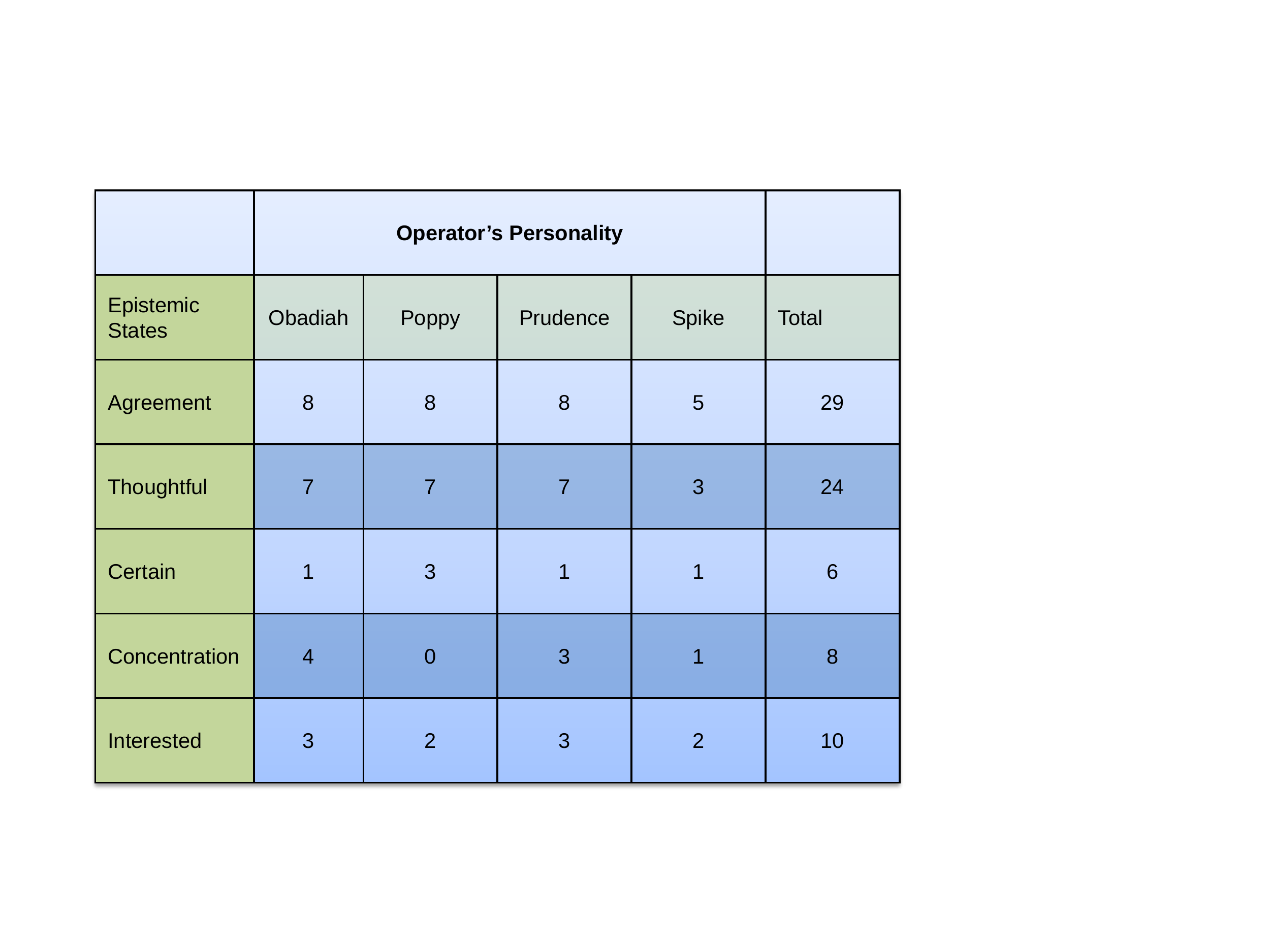}
\caption{The distributions of the video clips from operator's four types of characteristics}
\label{NumberofVideos}
\end{figure}

There are approximately 10 hours of footage available in the SAL database. For our detail analysis of five epistemic states, we have used corresponding videos from different characteristics. The figure \ref{NumberofVideos} shows the distributions of the video clips from operator's four types of characteristics.

\section{Experimental Results}
\label{sec:expResults}
Experiment results have two sections. The first one provides the observation  found  in the different experiments.  The second section shows the efficacy of the proposed approach for automated estimation of Epistemic Mental States and their intensities.
\subsection{Observations and Key Findings}
\label{subsec:Observations}
The key observations are listed below.

\begin{enumerate}
\item	Head roll, lip corner distance, and outer lip height are found to be highly correlated (non-linear) with concentration rating.
\item	First order temporal features (derived features from original facial features) correlate strongly with the changes of particular concentration rating.
\item	Duration of the epistemic state(concentration) is much longer than the prototypical one (anger). The window size 40 is optimal for concentration whereas window size six is good for anger.
\item The continuous ratings of concentration can be predicted robustly using two steps.
\begin{enumerate}
\item Classify the emotion changing regions (rising, falling, and steady regions) with original and  first-order temporal features. 
\item  Use three regression models for RISE, SUSTAIN, and DECAY regions. Use original features along with first-order temporal features for intensity prediction in rising and falling region.  Original features are robust enough in steady regions.
\end{enumerate}
\item	Certain and agreement are found to be strongly related.  Interest and concentration follows the same route.
\end{enumerate}

\subsection{Performance of Automatic Estimation}
\label{subsec:performance}

\begin{figure}[!h]
\centering
\includegraphics[ width = 3.45 in]{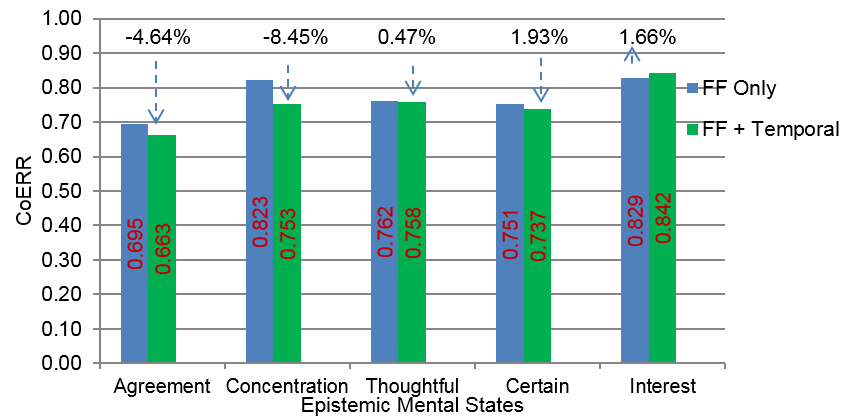}
\caption{Prediction result (Facial Features (FF) vs Temporal dynamics without any region information) - 10-fold cross validation on SEMAINE dataset videos}
\label{EpistemicMentalStatesAccuracy_WithoutRegion}
\end{figure}

After creating the framework for each of the mental states, all of the video sequences  from SEMAINE dataset are used to predict intensities frame by frame for corresponding EMS. For each of the EMS, we use separate models. To show the efficacy of our proposed models, we perform 10-fold cross validation on the SEMAINE dataset videos. Figure \ref{EpistemicMentalStatesAccuracy_WithoutRegion} shows that the use of facial features (FF) and  temporal features directly reduces the prediction performance except for Interest. The results support our analyses done in subsection \ref{subsec:regions}. It is easy notice that the use of temporal features do not improve the performance at all. Rather they make the performance worse. Concentration intensity prediction suffers mostly for the inclusion of temporal features. Correlation Coefficient (CoERR) between ground-truth intensity and predicted Concentration intensity is decreased by $8.45\%$. CoERR for Agreement is declined by $4.64\%$. With exception, CoERR for Interest is increased by $1.66\%$.
 
\begin{figure}[!h]
\centering
\includegraphics[ width = 3.45 in]{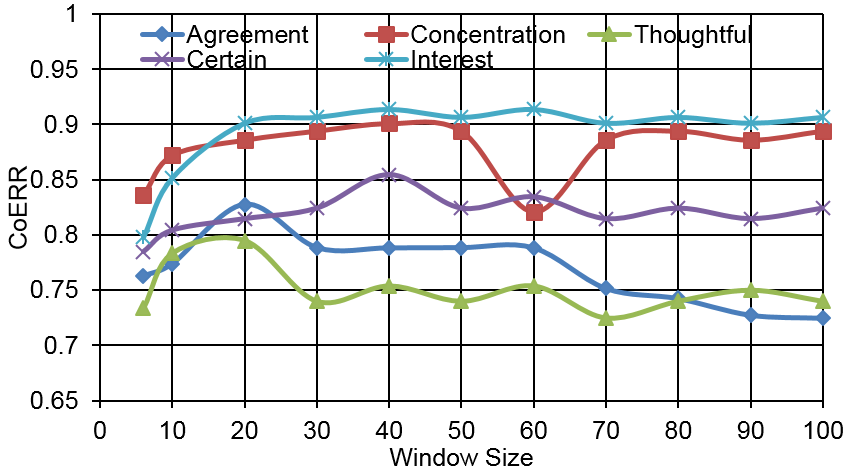}
\caption{Prediction accuracy of Epistemic States using varying window size (10-fold cross validation on SEMAINE dataset videos)}
\label{accuracyvsWindowsEpistemic}
\end{figure}

 Additionally we use our proposed framework for predicting the corresponding EMS  intensity. We use separate model for each EMS where region classifier is used in the intermediate stage as described in subsection \ref{subsec:Automated}. In these experiments, we use window size 20, 40, 20, 40, and 40 for the models of Agreement, Concentration, Thoughtful, Certain, and Interest, respectively. We select these particular window sizes based on the figure \ref{accuracyvsWindowsEpistemic}.
 
 The CoERR  increases by $19.06\%$ when we
 use our proposed model for Agreement. This is the most significant improvement among other EMSs. Certain's performance increases by $11.03\%$. Similarly, CoERR for Interest is improved by $10.25 \%$, Concentration by $8.67\%$,  and Thoughtful by $4.35\%$.  As observed in the table \ref{Nonlinearrelation},  yaw (left/right), pitch (up/down), and Lip Corner Distance  with strongest relations (MIC scores of 0.728, 0.683, and 0.675, respectively) with agreement state help accurate prediction of Agreement intensity. Similarly, for Certain,  Pitch, Lip Corner Distance, and Inner Brow with  high MIC scores (features weights) of $0.839$, $0.818$, and $0.760$ respectively  allow the proposed  model to improve the  prediction performance by $11.03 \%$. Analogous observations can be noticed for other EMSs and the corresponding facial features.

\begin{figure}[!h]
\centering
\includegraphics[ width = 3.45 in]{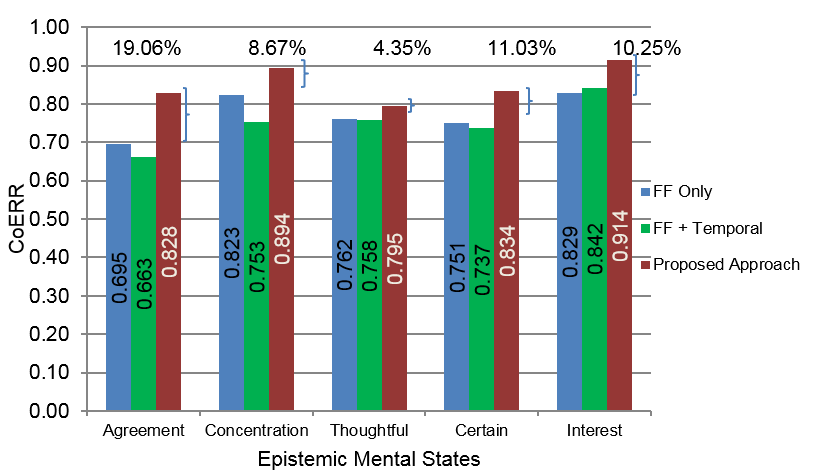}
\caption{Improvement of the prediction accuracy using the proposed approach (10-fold cross validation on SEMAINE dataset videos)}
\label{EpistemicMentalStatesAccuracy_WithRegion}
\end{figure}

We have performed another experiment to find how the EMSs evolve over time in SEMAINE videos. We have varied the window size from 5 to 100 and performed 10-fold cross validation. 
The figure \ref{accuracyvsWindowsEpistemic} shows that the accuracy of predicting epistemic mental states is varying over different window size. Whereas the categorical ones can be predicted using only window size 6 \cite{Bartlett06fullyautomatic}\cite{cohn2001recognizing}.

From the figure \ref{accuracyvsWindowsEpistemic}, it can be easily noticed that different epistemic mental states require different window size to model them robustly. For example, Concentration needs a window of size 40  to be modeled robustly. In contrast, Agreement and Thoughtful can be captured with window size 20. Interest can be equitably captured using 40 or 60 frames per window.

\section{Conclusion}
\label{sec:conc}
In this paper, we  propose novel methods to extract the non-linear relations between facial features and epistemic mental states: Agreement, Concentration, Thoughtful, Certain, and Interest. We have performed a number of   statistical analyses and simulations  to quantify these  relationships using maximal information coefficient. Particularly, non-linear relations are found to be more prevalent while temporal features derived from original facial features have demonstrated a strong correlation with intensity changes. Then, we propose a novel prediction framework that seems accurate in predicting epistemic state intensities. We propose to use the nonlinear relation-score (maximal information coefficient: MIC score) as weights of the facial features and their temporal dynamics. Moreover, the prediction of epistemic states is improved when the classification of emotion changing regions such as rising, falling, or sustaining are incorporated with the temporal features. In the experiments, the proposed predictive models are found to predict the epistemic states  with significantly improved accuracy: correlation coefficient (CoERR) for Agreement is $0.827$, for Concentration $0.901$, for Thoughtful $0.794$, for Certain $0.854$, and for Interest $0.913$. 
\\
\\
\textbf{Discussions and future works:}\\
The proposed framework lacks other modalities such as prosodic features, intonation patterns,  and  transcript of the conversations. Our work only includes the facial features to infer the EMSs. As discussed in the research context section, undoubtedly,  combination of all three  modalities ( verbal features  \cite{inproceedingsLinguistic}, prosodic features \cite{forbes2004predicting}, intonation patterns \cite{lee2002combining}, and facial features) would increase the EMS prediction performance. In future,  we are looking forward to including verbal and prosodic analysis to build a multimodal framework for EMSs prediction.

{
\bibliographystyle{refbib}
\bibliography{MTAP2020}

\begin{thebibliography}{10}

\bibitem{Mahmoud2011}
Marwa Mahmoud, Tadas Baltru\v{s}aitis, Peter Robinson, and Laurel Riek,
\newblock ``3d corpus of spontaneous complex mental states,''
\newblock in {\em Conference on Affective Computing and Intelligent
  Interaction}, 2011.

\bibitem{palm2013towards}
G{\"u}nther Palm and Michael Glodek,
\newblock ``Towards emotion recognition in human computer interaction,''
\newblock in {\em Neural Nets and Surroundings}, pp. 323--336. Springer, 2013.

\bibitem{shazia}
Shazia Afzal and Peter Robinson,
\newblock ``Modelling affect in learning environments - motivation and
  methods,''
\newblock in {\em ICALT '10: Proceedings of the 2010 10th IEEE International
  Conference on Advanced Learning Technologies}, 2010, pp. 438--442.

\bibitem{baron2007mind}
Simon Baron-Cohen,
\newblock ``Mind reading: the interactive guide to emotions--version 1.3,''
\newblock {\em Jessica Kingsley, London}, 2007.

\bibitem{semain}
Gary McKeown, Michel Valstar, Roddy Cowie, Maja Pantic, and Marc Schroder,
\newblock ``The semaine database: Annotated multimodal records of emotionally
  colored conversations between a person and a limited agent,''
\newblock {\em IEEE Trans. on Affective Computing}, vol. 3, pp. 5--17, 2012.

\bibitem{Roberts_Tsai_Coan_2007}
``Emotion elicitation using dyadic interaction tasks.,''
\newblock .

\bibitem{craig2004emotions}
Scotty Craig, Sidney D’Mello, Barry Gholson, Amy Witherspoon, Jeremiah
  Sullins, and Arthur Graesser,
\newblock ``Emotions during learning: The first steps toward an affect
  sensitive intelligent tutoring system,''
\newblock in {\em E-Learn: World Conference on E-Learning in Corporate,
  Government, Healthcare, and Higher Education}. Association for the
  Advancement of Computing in Education (AACE), 2004, pp. 264--268.

\bibitem{goleman1995emotional}
Daniel Goleman,
\newblock ``Emotional intelligence/d. golenman,''
\newblock {\em NY: Bantam Books}, 1995.

\bibitem{carterette1996handbook}
Edward~C Carterette, Morton~P Friedman, Joanne~L Miller, and Peter~D Eimas,
\newblock {\em Handbook of perception and cognition},
\newblock Academic Press New York, 1996.

\bibitem{mandler1975mind}
George Mandler,
\newblock {\em Mind and emotion},
\newblock Krieger Publishing Company, 1975.

\bibitem{parkinson1993making}
Brian Parkinson and Antony~SR Manstead,
\newblock ``Making sense of emotion in stories and social life,''
\newblock {\em Cognition \& Emotion}, vol. 7, no. 3-4, pp. 295--323, 1993.

\bibitem{kort2001affective}
Barry Kort, Rob Reilly, and Rosalind~W Picard,
\newblock ``An affective model of interplay between emotions and learning:
  Reengineering educational pedagogy-building a learning companion,''
\newblock in {\em Proceedings IEEE International Conference on Advanced
  Learning Technologies}. IEEE, 2001, pp. 43--46.

\bibitem{d2012autotutor}
Sidney D'mello and Art Graesser,
\newblock ``Autotutor and affective autotutor: Learning by talking with
  cognitively and emotionally intelligent computers that talk back,''
\newblock {\em ACM Transactions on Interactive Intelligent Systems (TiiS)},
  vol. 2, no. 4, pp. 23, 2012.

\bibitem{azevedo2009metatutor}
Roger Azevedo, Amy Witherspoon, Amber Chauncey, Candice Burkett, and Ashley
  Fike,
\newblock ``Metatutor: A metacognitive tool for enhancing self-regulated
  learning,''
\newblock in {\em 2009 AAAI Fall Symposium Series}, 2009.

\bibitem{breazeal2003emotion}
Cynthia Breazeal,
\newblock ``Emotion and sociable humanoid robots,''
\newblock {\em International journal of human-computer studies}, vol. 59, no.
  1-2, pp. 119--155, 2003.

\bibitem{borg2015assistive}
J~Borg, R~Berman-Bieler, C~Khasnabis, et~al.,
\newblock ``Assistive technology for children with disabilities: creating
  opportunities for education, inclusion and participation--a discussion
  paper,''
\newblock {\em Geneva: WHO}, 2015.

\bibitem{rahman2017emoassist}
AKMMahbubur Rahman, ASM~Iftekhar Anam, and Mohammed Yeasin,
\newblock ``Emoassist: emotion enabled assistive tool to enhance dyadic
  conversation for the blind,''
\newblock {\em Multimedia Tools and Applications}, vol. 76, no. 6, pp.
  7699--7730, 2017.

\bibitem{mcdaniel2018tactile}
Troy McDaniel, Diep Tran, Samjhana Devkota, Kaitlyn DiLorenzo, Bijan Fakhri,
  and Sethuraman Panchanathan,
\newblock ``Tactile facial expressions and associated emotions toward
  accessible social interactions for individuals who are blind,''
\newblock in {\em Proceedings of the 2018 Workshop on Multimedia for Accessible
  Human Computer Interface}. ACM, 2018, pp. 25--32.

\bibitem{Cheon2009}
Yeongjae Cheon and Daijin Kim,
\newblock ``Natural facial expression recognition using differential-aam and
  manifold learning,''
\newblock {\em Pattern Recognition}, vol. 42, no. 7, pp. 1340 -- 1350, 2009.

\bibitem{Zeng2007}
Zhihong Zeng, Maja Pantic, Glenn~I. Roisman, and Thomas~S. Huang,
\newblock ``A survey of affect recognition methods: audio, visual and
  spontaneous expressions,''
\newblock pp. 126--133, 2007.

\bibitem{sydneystudy07}
B~McDaniel, S~DMello, B~King, P~Chipman, K~Tapp, and A~Graesser,
\newblock ``Facial features for affective state detection in learning
  environments,''
\newblock in {\em 29th Annual Cognitive Science Society}. 2007, pp. 467--472,
  Cognitive Science Society.

\bibitem{surveysppechchannel}
Zhihong Zeng, M.~Pantic, G.I. Roisman, and T.S. Huang,
\newblock ``A survey of affect recognition methods: Audio, visual, and
  spontaneous expressions,''
\newblock {\em IEEE Trans. on PAMI}, vol. 31, no. 1, pp. 39 --58, 2009.

\bibitem{huang2007labeled}
Gary~B Huang, Marwan Mattar, Tamara Berg, and Erik Learned-Miller,
\newblock ``E.: Labeled faces in the wild: A database for studying face
  recognition in unconstrained environments,''
\newblock 2007.

\bibitem{gordon2016affective}
Goren Gordon, Samuel Spaulding, Jacqueline~Kory Westlund, Jin~Joo Lee, Luke
  Plummer, Marayna Martinez, Madhurima Das, and Cynthia Breazeal,
\newblock ``Affective personalization of a social robot tutor for children’s
  second language skills,''
\newblock in {\em Thirtieth AAAI Conference on Artificial Intelligence}, 2016.

\bibitem{cohn_schemidt}
Jeffrey~F. Cohn, Karen Schmidt, Ralph Gross, and Paul Ekman,
\newblock ``Individual differences in facial expression: Stability over ti
  relation to self-reported emotion, and ability to inform person
  identification,''
\newblock in {\em 4th IEEE ICME}, Washington, DC, USA, 2002, ICMI '02, pp.
  491--, IEEE Computer Society.

\bibitem{mcdaniel2007facial1}
Bethany McDaniel, Sidney D'Mello, Brandon King, Patrick Chipman, Kristy Tapp,
  and Art Graesser,
\newblock ``Facial features for affective state detection in learning
  environments,''
\newblock in {\em Proceedings of the Annual Meeting of the Cognitive Science
  Society}, 2007, vol.~29.

\bibitem{bosch2014s}
Nigel Bosch, Yuxuan Chen, and Sidney D’Mello,
\newblock ``It’s written on your face: detecting affective states from facial
  expressions while learning computer programming,''
\newblock in {\em International Conference on Intelligent Tutoring Systems}.
  Springer, 2014, pp. 39--44.

\bibitem{littlewort2011computer}
Gwen Littlewort, Jacob Whitehill, Tingfan Wu, Ian Fasel, Mark Frank, Javier
  Movellan, and Marian Bartlett,
\newblock ``The computer expression recognition toolbox (cert),''
\newblock in {\em Face and gesture 2011}. IEEE, 2011, pp. 298--305.

\bibitem{bosch2016detecting}
Nigel Bosch, Sidney~K D'Mello, Ryan~S Baker, Jaclyn Ocumpaugh, Valerie Shute,
  Matthew Ventura, Lubin Wang, and Weinan Zhao,
\newblock ``Detecting student emotions in computer-enabled classrooms.,''
\newblock in {\em IJCAI}, 2016, pp. 4125--4129.

\bibitem{reshef2011detecting}
David~N Reshef, Yakir~A Reshef, Hilary~K Finucane, Sharon~R Grossman, Gilean
  McVean, Peter~J Turnbaugh, Eric~S Lander, Michael Mitzenmacher, and Pardis~C
  Sabeti,
\newblock ``Detecting novel associations in large data sets,''
\newblock {\em science}, vol. 334, no. 6062, pp. 1518--1524, 2011.

\bibitem{d2010multimodal}
Sidney~K D?Mello and Arthur Graesser,
\newblock ``Multimodal semi-automated affect detection from conversational
  cues, gross body language, and facial features,''
\newblock {\em User Modeling and User-Adapted Interaction}, vol. 20, no. 2, pp.
  147--187, 2010.

\bibitem{metatutor}
Vincent Aleven, Bruce Mclaren, Ido Roll, and Kenneth Koedinger,
\newblock ``Toward meta-cognitive tutoring: A model of help seeking with a
  cognitive tutor,''
\newblock {\em Int. J. Artif. Intell. Ed.}, vol. 16, no. 2, pp. 101--128, 2006.

\bibitem{Graesser_mindand}
Sidney D'mello and Arthur Graesser,
\newblock ``Mind and body: Dialogue and posture for affect detection in
  learning environments,''
\newblock in {\em 2007 conference on Artificial Intelligence in Education},
  Amsterdam, The Netherlands, The Netherlands, 2007, pp. 161--168, IOS Press.

\bibitem{lanzini2013different}
Stefano Lanzini,
\newblock ``How do different modes contribute to the interpretation of
  affective epistemic states,''
\newblock {\em Published master’s thesis for master’s degree. University
  Gothenburg, Division of Communication and Cognition, Department of Applied
  IT}, 2013.

\bibitem{knapp2013nonverbal}
Mark~L Knapp, Judith~A Hall, and Terrence~G Horgan,
\newblock {\em Nonverbal communication in human interaction},
\newblock Cengage Learning, 2013.

\bibitem{cohn2010advances}
Jeffrey~F Cohn,
\newblock ``Advances in behavioral science using automated facial image
  analysis and synthesis [social sciences],''
\newblock {\em IEEE Signal processing magazine}, vol. 27, no. 6, pp. 128--133,
  2010.

\bibitem{stratou2017refactoring}
Giota Stratou, Job Van Der~Schalk, Rens Hoegen, and Jonathan Gratch,
\newblock ``Refactoring facial expressions: An automatic analysis of natural
  occurring facial expressions in iterative social dilemma,''
\newblock in {\em 2017 Seventh International Conference on Affective Computing
  and Intelligent Interaction (ACII)}. IEEE, 2017, pp. 427--433.

\bibitem{littlewort2011automated}
Gwen~C Littlewort, Marian~S Bartlett, Linda~P Salamanca, and Judy Reilly,
\newblock ``Automated measurement of children's facial expressions during
  problem solving tasks,''
\newblock in {\em Face and Gesture 2011}. IEEE, 2011, pp. 30--35.

\bibitem{jackson2009enhanced}
Margaret~C Jackson, Chia-Yun Wu, David~EJ Linden, and Jane~E Raymond,
\newblock ``Enhanced visual short-term memory for angry faces.,''
\newblock {\em Journal of Experimental Psychology: Human Perception and
  Performance}, vol. 35, no. 2, pp. 363, 2009.

\bibitem{dimberg2000unconscious}
Ulf Dimberg, Monika Thunberg, and Kurt Elmehed,
\newblock ``Unconscious facial reactions to emotional facial expressions,''
\newblock {\em Psychological science}, vol. 11, no. 1, pp. 86--89, 2000.

\bibitem{schneider1992preschoolers}
Klaus Schneider and Lothar Unzner,
\newblock ``Preschoolers' attention and emotion in an achievement and an effect
  game: A longitudinal study,''
\newblock {\em Cognition \& Emotion}, vol. 6, no. 1, pp. 37--63, 1992.

\bibitem{autotutor}
Sidney D'Mello, Rosalind~W. Picard, and Arthur Graesser,
\newblock ``Toward an affect-sensitive autotutor,''
\newblock {\em IEEE Intelligent Systems}, vol. 22, pp. 53--61, 2007.

\bibitem{hoque2013mach}
Mohammed~Ehsan Hoque, Matthieu Courgeon, Jean-Claude Martin, Bilge Mutlu, and
  Rosalind~W Picard,
\newblock ``Mach: My automated conversation coach,''
\newblock in {\em Proceedings of the 2013 ACM international joint conference on
  Pervasive and ubiquitous computing}. ACM, 2013, pp. 697--706.

\bibitem{bousmalis2013towards}
Konstantinos Bousmalis, Marc Mehu, and Maja Pantic,
\newblock ``Towards the automatic detection of spontaneous agreement and
  disagreement based on nonverbal behaviour: A survey of related cues,
  databases, and tools,''
\newblock {\em Image and Vision Computing}, vol. 31, no. 2, pp. 203--221, 2013.

\bibitem{mehu2014multimodal}
Marc Mehu and Laurens Van~der Maaten,
\newblock ``Multimodal integration of dynamic audio--visual cues in the
  communication of agreement and disagreement,''
\newblock {\em Journal of Nonverbal Behavior}, vol. 38, no. 4, pp. 569--597,
  2014.

\bibitem{poggi2010agreement}
I~Poggi, F~D’Errico, and L~Vincze,
\newblock ``Agreement and its multimodal communication in debates,''
\newblock {\em A qualitative analysis. Cognitive Computation. doi}, vol. 10,
  2010.

\bibitem{vinciarelli2009canal9}
Alessandro Vinciarelli, Alfred Dielmann, Sarah Favre, and Hugues Salamin,
\newblock ``Canal9: A database of political debates for analysis of social
  interactions,''
\newblock in {\em 2009 3rd International Conference on Affective Computing and
  Intelligent Interaction and Workshops}. IEEE, 2009, pp. 1--4.

\bibitem{carletta2003nite}
Jean Carletta, Stefan Evert, Ulrich Heid, Jonathan Kilgour, Judy Robertson, and
  Holger Voormann,
\newblock ``The nite xml toolkit: flexible annotation for multimodal language
  data,''
\newblock {\em Behavior Research Methods, Instruments, \& Computers}, vol. 35,
  no. 3, pp. 353--363, 2003.

\bibitem{pachman2016eye}
Mariya Pachman, Ama{\"e}l Arguel, Lori Lockyer, Gregor Kennedy, and Jason
  Lodge,
\newblock ``Eye tracking and early detection of confusion in digital learning
  environments: Proof of concept,''
\newblock {\em Australasian Journal of Educational Technology}, vol. 32, no. 6,
  2016.

\bibitem{grafsgaard2011modeling}
Joseph~F Grafsgaard, Kristy~Elizabeth Boyer, Robert Phillips, and James~C
  Lester,
\newblock ``Modeling confusion: facial expression, task, and discourse in
  task-oriented tutorial dialogue,''
\newblock in {\em International Conference on Artificial Intelligence in
  Education}. Springer, 2011, pp. 98--105.

\bibitem{krithika2016student}
LB~Krithika,
\newblock ``Student emotion recognition system (sers) for e-learning
  improvement based on learner concentration metric,''
\newblock {\em Procedia Computer Science}, vol. 85, pp. 767--776, 2016.

\bibitem{d2018cognitive}
Francesca D'Errico, Marinella Paciello, Bernardina De~Carolis, Alessandro
  Vattanid, Giuseppe Palestra, and Giuseppe Anzivino,
\newblock ``Cognitive emotions in e-learning processes and their potential
  relationship with students’ academic adjustment,''
\newblock 2018.

\bibitem{de2019engaged}
Berardina De~Carolis, Francesca D'Errico, Nicola Macchiarulo, and Giuseppe
  Palestra,
\newblock ``“engaged faces”: Measuring and monitoring student engagement
  from face and gaze behavior,''
\newblock in {\em IEEE/WIC/ACM International Conference on Web
  Intelligence-Companion Volume}, 2019, pp. 80--85.

\bibitem{d2016emotions}
Francesca D'Errico, Marinella Paciello, and Luca Cerniglia,
\newblock ``When emotions enhance students’ engagement in e-learning
  processes,''
\newblock {\em Journal of e-Learning and Knowledge Society}, vol. 12, no. 4,
  2016.

\bibitem{hung2010estimating}
Hayley Hung and Daniel Gatica-Perez,
\newblock ``Estimating cohesion in small groups using audio-visual nonverbal
  behavior,''
\newblock {\em Multimedia, IEEE Transactions on}, vol. 12, no. 6, pp. 563--575,
  2010.

\bibitem{nguyen2012using}
Laurent Nguyen, Jean-Marc Odobez, and Daniel Gatica-Perez,
\newblock ``Using self-context for multimodal detection of head nods in
  face-to-face interactions,''
\newblock in {\em Proceedings of the 14th ACM international conference on
  Multimodal interaction}. ACM, 2012, pp. 289--292.

\bibitem{arango2014nature}
Santiago Arango-Mu{\~n}oz,
\newblock ``The nature of epistemic feelings,''
\newblock {\em Philosophical Psychology}, vol. 27, no. 2, pp. 193--211, 2014.

\bibitem{michaelian2014epistemic}
Kourken Michaelian and Santiago Arango-Mu{\~A}$\pm$oz,
\newblock ``Epistemic feelings, epistemic emotions: Review and introduction to
  the focus section,''
\newblock {\em Philosophical inquiries}, vol. 2, no. 1, pp. 97--122, 2014.

\bibitem{inproceedingsLinguistic}
Laurence Devillers and Laurence Vidrascu,
\newblock ``Real-life emotions detection with lexical and paralinguistic cues
  on human-human call center dialogs.,''
\newblock 01 2006.

\bibitem{forbes2004predicting}
Kate Forbes-Riley and Diane Litman,
\newblock ``Predicting emotion in spoken dialogue from multiple knowledge
  sources,''
\newblock in {\em Proceedings of the Human Language Technology Conference of
  the North American Chapter of the Association for Computational Linguistics:
  HLT-NAACL 2004}, 2004, pp. 201--208.

\bibitem{lee2002combining}
Chul~Min Lee, Shrikanth~S Narayanan, and Roberto Pieraccini,
\newblock ``Combining acoustic and language information for emotion
  recognition,''
\newblock in {\em Seventh International Conference on Spoken Language
  Processing}, 2002.

\bibitem{borras2011perceiving}
Joan Borr{\`a}s-Comes, Paolo Roseano, Maria del~Mar Vanrell, Aoju Chen, and
  Pilar Prieto,
\newblock ``Perceiving uncertainty: facial gestures, intonation, and lexical
  choice,''
\newblock {\em Proceedings of GESPIN}, 2011.

\bibitem{contVA}
M.A. Nicolaou, H.~Gunes, and M.~Pantic,
\newblock ``Continuous prediction of spontaneous affect from multiple cues and
  modalities in valence-arousal space,''
\newblock {\em Affective Computing, IEEE Trans. on}, vol. 2, no. 2, 2011.

\bibitem{lambDom}
Theodore~A. Lamb,
\newblock ``Nonverbal and paraverbal control in dyads and triads: Sex or power
  differences?,''
\newblock {\em Social Psychology Quarterly}, vol. 44, no. 1, pp. pp. 49--53,
  1981.

\bibitem{dominanceEyeGaze}
N.~Bee, S.~Franke, and E.~Andrea,
\newblock ``Relations between facial display, eye gaze and head tilt: Dominance
  perception variations of virtual agents,''
\newblock in {\em ACII Workshop 2009}, sept. 2009.

\bibitem{Saragih3}
J.M. Saragih, S.~Lucey, and J.F. Cohn,
\newblock ``Face alignment through subspace constrained mean-shifts,''
\newblock in {\em Computer Vision, 2009 IEEE 12th International Conference on}.
  Ieee, 2009, pp. 1034--1041.

\bibitem{atkinson2005visual}
A.P. ATKINSON and R.~ADOLPHS,
\newblock ``Visual emotion perception: Mechanisms and processes.,''
\newblock {\em Emotion and consciousness}, p. 150, 2005.

\bibitem{reviewJournal}
Z.~Zeng, M.~Pantic, G.I. Roisman, and T.S. Huang,
\newblock ``A survey of affect recognition methods: Audio, visual, and
  spontaneous expressions,''
\newblock {\em Pattern Analysis and Machine Intelligence, IEEE Transactions
  on}, vol. 31, no. 1, pp. 39--58, Jan.

\bibitem{da2009epistemic}
Ronald da~Sousa,
\newblock ``Epistemic feelings,''
\newblock {\em Mind and Matter}, vol. 7, no. 2, pp. 139--161, 2009.

\bibitem{vapnik1997support}
Vladimir Vapnik, Steven~E Golowich, and Alex Smola,
\newblock ``Support vector method for function approximation, regression
  estimation, and signal processing,''
\newblock {\em Advances in neural information processing systems}, pp.
  281--287, 1997.

\bibitem{tomkins1962affect}
Silvan~S Tomkins,
\newblock ``Affect, imagery, consciousness: Vol. i. the positive affects.,''
\newblock 1962.

\bibitem{Bartlett06fullyautomatic}
M~S Bartlett, G~Littlewort, M~G Frank, C~Lainscsek, I~Fasel, and J~Movellan,
\newblock ``Fully automatic facial action recognition in spontaneous
  behavior,''
\newblock {\em Journal of Multimedia}, , no. 6, pp. 22--35, 2006.

\bibitem{cohn2001recognizing}
Ying-liTian TakeoKanade~JeffreyF Cohn,
\newblock ``Recognizing facial actions by combining geometric features and
  regional appearance patterns,''
\newblock 2001.

\end{thebibliography}
}

\end{document}